\documentclass{article} 
\usepackage[final]{colm2026_conference}

\usepackage[T1]{fontenc}
\usepackage{microtype}
\usepackage{booktabs}
\usepackage{graphicx}   
\usepackage{multirow}
\usepackage{amsmath}
\usepackage{xcolor}
\usepackage{colortbl}
\ifcolmsubmission\usepackage{lineno}\fi
\usepackage{float}      
\usepackage{placeins}   

\PassOptionsToPackage{hyphens}{url}
\usepackage{hyperref}
\usepackage{url}

\definecolor{darkblue}{rgb}{0, 0, 0.5}
\hypersetup{colorlinks=true, citecolor=darkblue, linkcolor=darkblue, urlcolor=darkblue}

\graphicspath{{figures/}}

\title{The Embedder's Dilemma: LLMs Are Better, \\but at What Cost?}

\author{Adnan El Assadi \\
Harvard University
\And
Niklas Muennighoff \\
Stanford University
\And
Jinhyuk Lee \\
Independent Researcher}

\newcommand{\mteblm}{MTEB(LLM)}

\begin{document}

\setlength{\textfloatsep}{8pt plus 2pt minus 2pt}
\setlength{\floatsep}{8pt plus 2pt minus 2pt}
\setlength{\abovecaptionskip}{6pt}
\setlength{\belowcaptionskip}{2pt}

\ifcolmsubmission
\linenumbers
\fi

\maketitle

\begin{abstract}
Should you replace your text-embedding pipeline with a large language model?
We answer this with a controlled, cost-aware comparison of ten LLMs across six families and 26 embedding models (118M to 14B parameters) on 37 tasks spanning classification, semantic textual similarity (STS), clustering, pair classification, and retrieval.
In aggregate the two paradigms are effectively tied: the best LLM (Gemini~3.1~Pro, 77.6) and the best embedding model (77.2) differ by 0.4 points.
Their strengths differ by task: LLMs lead on reasoning-heavy retrieval, embedding models lead on classification, and the two match on clustering, STS, and pair classification.
Reaching that parity is expensive. An LLM costs up to 1{,}431$\times$ more than an embedding model of comparable quality (\$154 vs.\ \$0.11 per benchmark pass), and the open LLMs tested process tokens 2.5 to 736$\times$ more slowly on the same GPU.
Reasoning tokens account for 28 to 81\% of LLM inference cost; lower reasoning budgets preserve or improve retrieval quality for most models in our ablation.
The Pareto frontier contains the leading embedding models and one LLM, Gemini~3.1~Pro.
These results support a division of labour: use embedding models for similarity, classification, and clustering, and reserve LLMs for reasoning-intensive retrieval.
Our code, datasets, and results are publicly available at \url{https://github.com/embeddings-benchmark/embedders-dilemma}.
\end{abstract}

\begin{figure}[H]
\centering
\includegraphics[width=\textwidth]{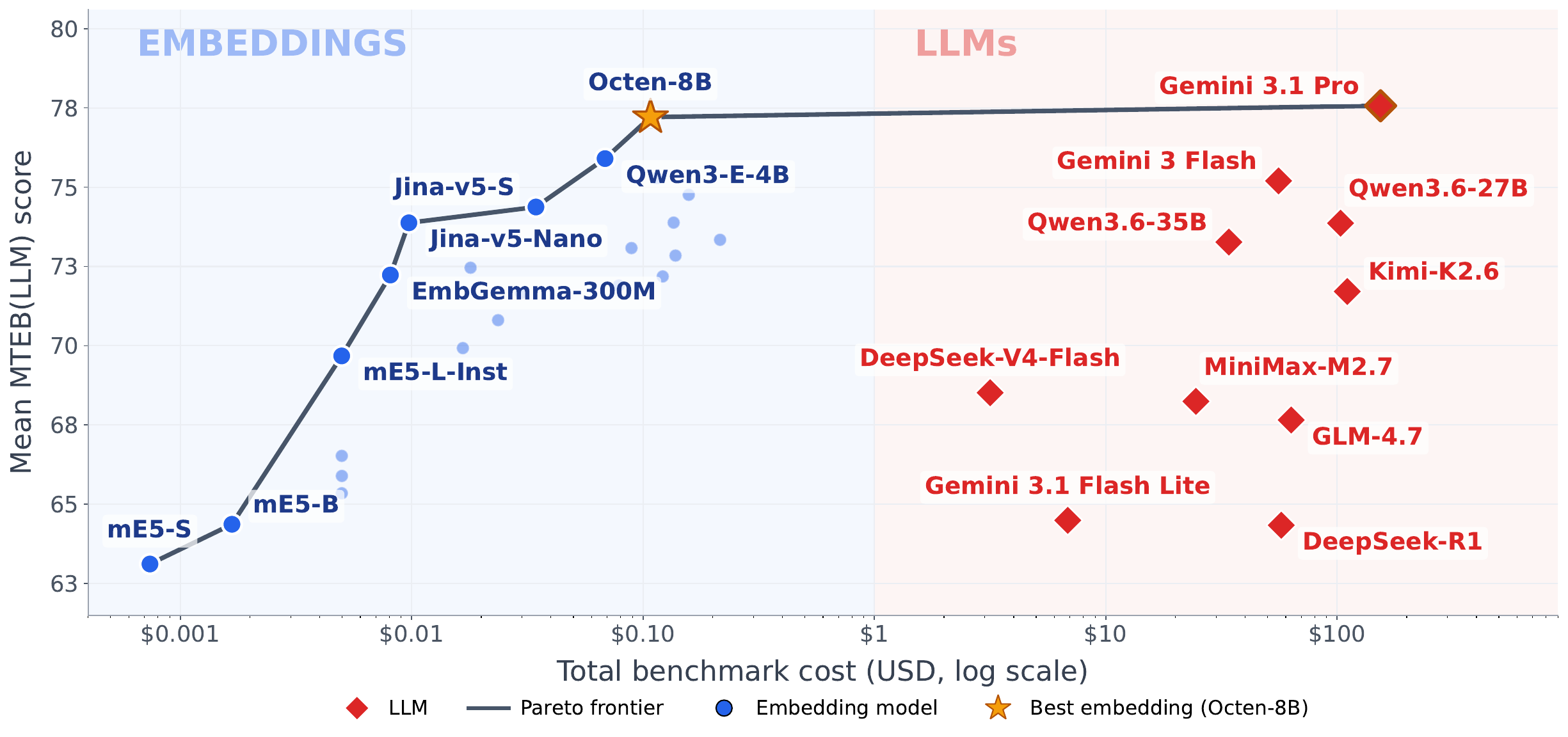}
\caption{\textbf{Cost vs.\ performance across 36 models on \mteblm.}
The frontier contains the leading embedding models and Gemini~3.1~Pro, which extends it by 0.4 points at 1{,}431$\times$ the cost of a comparable embedding.}
\label{fig:pareto}
\end{figure}

\section{Introduction}
\label{sec:intro}

We show that an LLM without task-specific training now matches the best text embedding models across a broad suite of standard embedding tasks. 
Embedding models reach this quality through specialised contrastive training, hard-negative mining, and multi-stage distillation; our best LLM reaches it without a dedicated embedding-training pipeline.\footnote{While embedding models have increasingly adopted LLMs as their backbone~\citep{tao2024llms,nie2025textembeddingmeetslarge}, they output vectors and use training tailored to embedding tasks. Meanwhile, ``LLMs'' in this work are generative models prompted to output text, with no dedicated embedding training.}
The deployment costs differ sharply: LLM inference is substantially more expensive and slower on the same hardware.
We call this the \textbf{Embedder's Dilemma}: LLMs now match the best embedding models in aggregate, but at much higher cost and lower throughput.

We ask a practical question: \emph{should you replace your embedding pipeline with an LLM?}
Our results favour embedding models as the default and LLMs for reasoning-heavy retrieval. A hybrid pipeline captures the retrieval benefit while controlling cost and throughput.

We compare ten frontier LLMs from six families with 26 text embedding models (118M--14B parameters). The evaluation uses \textbf{\mteblm}, a 37-task benchmark built from the Massive Text Embedding Benchmark \citep[MTEB;][]{muennighoff2023mteb}, covering classification, semantic textual similarity (STS), clustering, pair classification, and retrieval. The full model list appears in \S\ref{sec:setup}.
Each \mteblm{} task is a fixed subset of the corresponding MTEB task, released on Hugging Face under \texttt{mteb/llm-eval-*}; both paradigms are scored on exactly that subset.
\mteblm{} scores compare models within this benchmark; MTEB leaderboard scores use the full test sets and follow a different evaluation basis.
We pair the quality comparison with exact cost accounting (API token tracking for LLMs; GPU throughput benchmarking for embedding models), producing a cost--performance Pareto frontier spanning both paradigms (Figure~\ref{fig:pareto}).

\textbf{Contributions.}
We release \mteblm{}, the first benchmark to compare LLMs and embedding models across all five MTEB task categories, with cost measured alongside quality for every model.
We implement it with the MTEB framework, part of an ecosystem that spans multilingual text (MMTEB; \citealp{enevoldsen2025mmtebmassivemultilingualtext}), image (MIEB; \citealp{xiao2025mieb}), audio (MAEB; \citealp{elassadi2026maeb}), and video (MVEB; \citealp{elassadi2026mveb}).
The Pareto frontier contains the leading embeddings and Gemini~3.1~Pro, whose small score gain comes at three orders of magnitude higher cost.
We identify the \emph{thinking-token tax}: reasoning accounts for 28--81\% of LLM inference cost, while lower reasoning budgets preserve retrieval quality for most models in our ablation (\S\ref{sec:ablation}).

Our empirical findings are:
\begin{enumerate}
    \item \textbf{In aggregate, the paradigms tie.}
    The best LLM and embedding model differ by 0.4 points, within statistical noise (\S\ref{sec:categories}).
    The pattern holds across the task suite.

    \item \textbf{The advantage is task-specific.}
    LLMs lead on reasoning-heavy retrieval, embedding models lead on classification, and the paradigms are statistically even on clustering, STS, and pair classification.
    LLMs excel when cross-document reasoning is required; embedding pipelines excel wherever geometric similarity or labelled-reference matching suffices (\S\ref{sec:tasks}--\ref{sec:categories}).

    \item \textbf{The cost gap is large.}
    Gemini~3.1~Pro costs 1{,}431$\times$ as much as a comparable embedding model, and the gap persists across hardware scenarios.
    Reasoning contributes substantially to this asymmetry. Lower reasoning budgets preserve or improve retrieval for most evaluated models (\S\ref{sec:cost_results}--\ref{sec:ablation}).

    \item \textbf{Throughput lags by orders of magnitude, even on the same hardware.}
    Served on an identical H100, open-weight LLMs process 2.5--736$\times$ fewer tokens per second than embedding models, consistent with the architectural difference between autoregressive generation and a single encoder pass (\S\ref{sec:throughput_results}; Figure~\ref{fig:throughput}).
\end{enumerate}

LLMs and embedding models therefore serve complementary roles. Embedding pipelines remain the efficient default for classification, similarity, and clustering; LLMs are most useful for reasoning-intensive retrieval.

\section{Related Work}
\label{sec:related}

\subsection{Text Embedding Models, Evaluation, and Cost}

Dense text representations have evolved rapidly from contextualised encoders \citep{devlin2019bert, reimers2019sentencebert} to billion-parameter, instruction-tuned models \citep{su2022instructor}.
Leading systems such as E5 \citep{wang2024textembeddingsweaklysupervisedcontrastive, wang2024improvingtextembeddingslarge}, NV-Embed \citep{lee2024nvembedimprovedtechniques}, SFR-2 \citep{meng2024sfremb2}, and Gemini Embedding \citep{lee2025geminiembedding} achieve strong performance through contrastive training on large curated corpora.
MTEB \citep{muennighoff2023mteb} standardised evaluation across eight task categories and now tracks hundreds of models on a public leaderboard.\footnote{\url{https://huggingface.co/spaces/mteb/leaderboard}}
The framework has expanded across multilingual text (MMTEB; \citealp{enevoldsen2025mmtebmassivemultilingualtext}), image (MIEB; \citealp{xiao2025mieb}), audio (MAEB; \citealp{elassadi2026maeb}), and video (MVEB; \citealp{elassadi2026mveb}).
HUME \citep{assadi2025humemeasuringhumanmodelperformance} measures human performance on MTEB tasks and benchmarks LLMs as annotators. Our study asks whether an LLM can replace an embedding pipeline.
We extend this line of evaluation with exact cost and throughput accounting across both paradigms.

\citet{schwartz2020green} introduced ``Green AI,'' arguing that accuracy focus neglects computational and environmental costs.
\citet{gonzalez2026costaware} constructs Pareto frontiers of fine-tuned BERT-scale encoders vs.\ GPT-4o on text classification, finding encoders match or exceed frontier LLMs at 1--2 orders of magnitude lower cost.
The ``overthinking'' literature \citep{sui2025stopoverthinking} documents verbose reasoning traces and their computational cost. \citet{snell2024scalingllmtesttimecompute} show that adaptive test-time compute can outperform a uniform reasoning budget. We study these questions on classification, similarity, clustering, and retrieval workloads.

\subsection{LLMs, Cross-Document Reasoning, and Retrieval}

LLMs with controllable reasoning now span closed APIs (Gemini~3.1~Pro, GPT-5 \citep{singh2026openaigpt5card}, Claude~Opus~4.6 \citep{anthropic2026claudeopus46}) and open weights (Qwen3 \citep{yang2025qwen3technicalreport}, Llama~3 \citep{grattafiori2024llama3herdmodels}, and DeepSeek-R1 \citep{deepseekai2025deepseekr1incentivizingreasoningcapability}, among others).
We evaluate ten models across six families (\S\ref{sec:setup}), including dense and mixture-of-experts models from closed and open providers. This diversity lets us distinguish recurring patterns from effects specific to one provider.
\citet{bucher2024finetuned} find that fine-tuned encoder models significantly outperform zero-shot frontier LLMs across all tested classification benchmarks, with margins ranging from 5 to over 75 F1 points depending on task granularity, consistent with our 5.6-point embedding advantage under the kNN vs.\ zero-shot comparison.

Dense retrieval \citep{karpukhin2020dpr} and the BEIR benchmark \citep{thakur2021beir} established strong zero-shot bi-encoder performance. Bi-encoders process queries and documents independently, without cross-attention between them.
The BRIGHT benchmark \citep{su2024bright} complements BEIR with reasoning-intensive queries. Its top embedding model scores 18.3 nDCG@10, while LLM-augmented retrieval improves by up to 12.2 points, showing the limitations of independent query and document encoding on such tasks.
RAG \citep{lewis2020rag, gao2024retrievalaugmentedgenerationlargelanguage} pairs embedding retrieval with LLM reasoning. We evaluate this reranking stage directly \citep{sun2023rankgpt}, crossing first-stage retrievers with cross-encoder and LLM listwise rerankers on BEIR and BRIGHT (\S\ref{sec:rerank}).
\citet{lu2025cothurts} show CoT reranking consistently underperforms direct-output reranking on BEIR/BRIGHT despite higher cost, the document-ranking counterpart to our finding that reduced thinking holds or improves retrieval on all six tasks.
Appendix~\ref{app:extended_related} covers LLMs as text encoders, GritLM, and the wider set of frontier models.

\section{Experimental Setup}
\label{sec:setup}

\subsection{Models}

\paragraph{LLMs.}
We evaluate ten frontier LLMs spanning six families: three Gemini~3 models via the Gemini API (Gemini~3.1~Pro, Gemini~3~Flash, Gemini~3.1~Flash-Lite) and seven open-weight models served via OpenRouter (DeepSeek-R1, DeepSeek-V4-Flash, Qwen3.6-27B, Qwen3.6-35B-A3B, GLM-4.7, Kimi-K2.6, and MiniMax-M2.7).
Together they span dense and mixture-of-experts architectures, reasoning and instruct variants, and a wide capability and cost range.
Gemini~3.1~Flash-Lite provides a non-reasoning baseline; the other models support extended reasoning (chain-of-thought tokens \citep{wei2022cot}, billed at the output token rate).

\paragraph{Embedding models.}
We evaluate 26 text embedding models from 118M to 14B parameters. The multi-model families are multilingual-E5 \citep{wang2024textembeddingsweaklysupervisedcontrastive,wang2024improvingtextembeddingslarge}, Qwen3-Embedding \citep{qwen2025qwen3embedding}, F2LLM-v2 \citep{zhang2026f2llmv2}, Jina-v5 \citep{akram2026jinaembeddingsv5}, and GTE-Qwen2 \citep{li2023generaltextembeddingsmultistage}. We also include models from Tencent \citep{hu2025kalmv2}, NVIDIA \citep{babakhin2025llamaembednemotron}, Salesforce \citep{meng2024sfremb2}, Snowflake \citep{yu2024arctic}, and Google \citep{vera2025embeddinggemma}, together with E5-Mistral \citep{wang2024improvingtextembeddingslarge}, GritLM \citep{muennighoff2025generativerepresentationalinstructiontuning}, Linq-Embed-Mistral \citep{choi2024linq}, BGE-M3 \citep{chen2025m3embeddingmultilingualitymultifunctionalitymultigranularity}, and Octen-8B.
All embedding models run locally on a single NVIDIA H100 80GB HBM3 GPU.
Table~\ref{tab:main_results} lists every model with its score and cost; full details appear in Appendix~\ref{app:models_detail}.

\subsection{Tasks and Evaluation Protocol}
\label{sec:tasks}

We introduce \textbf{\mteblm}, a 37-task benchmark covering classification, STS, clustering, pair classification, and retrieval, implemented with the MTEB framework \citep{muennighoff2023mteb}.
It follows MTEB's task and result interfaces, allowing new models and datasets to use the same evaluation pipeline.
Each \mteblm{} task is a new LLM-specific evaluation dataset derived from the corresponding original MTEB task (fixed seed\,=\,42; Appendix~\ref{app:datasets}), ensuring all submissions are evaluated on identical data.

\paragraph{Why these tasks.}
We use subsets rather than MTEB(eng, v2) because generative evaluation is orders of magnitude more expensive: one pass over the full suite would cost hundreds of dollars per LLM, and corpus-in-context retrieval requires the corpus to fit in the prompt.
Tasks are selected to span all five categories, cover multiple languages and domains, and keep retrieval corpora small enough for shorter-context models.
MTEB's task-selection analysis was also designed around embedding models, so v2 is not automatically the right frame for a cross-paradigm comparison.

We compare complete deployment pipelines: embedding models use kNN for classification, cosine similarity for STS and retrieval, and $k$-means for clustering; LLMs receive zero-shot prompts.
This comparison reflects a common deployment setting in which practitioners consider LLMs because labelled data are scarce.
Our few-shot ablation (\S\ref{sec:fewshot}) tests whether a small number of in-context examples narrows the resulting gap.

\paragraph{Classification (8 tasks).}
Sentiment (IMDB, ToxicConversations, TweetSentiment), intent detection (Banking77: 77 classes; MassiveIntent: 60 intents), and multilingual classification (AmazonCounterfactual, MTOPDomain, MassiveScenario).
\emph{LLM:} Zero-shot structured output for each test split sample.
\emph{Embedding:} kNN trained on embeddings on the labelled training split, then used for test split predictions.
\emph{Metric:} Accuracy.

\paragraph{Semantic Textual Similarity (10 tasks).}
English benchmarks (STSBenchmark, SICK-R, STS12--16), a biomedical benchmark (BIOSSES), and multilingual benchmarks (STS17, STS22v2).
\emph{LLM:} Rates similarity on each task's native scale, returning a float.
\emph{Embedding:} Cosine similarity between embeddings.
\emph{Metric:} Spearman correlation.

\paragraph{Clustering (9 tasks).}
Scientific papers (ArXiv, BioRxiv, MedRxiv) and online discussions (Reddit, StackExchange, TwentyNewsgroups).
\emph{LLM:} All documents in a single prompt with sequential identifiers; the model receives the ground-truth cluster count $k$ and returns cluster assignments as a JSON array.
\emph{Embedding:} $k$-means on document embeddings.
\emph{Metric:} V-measure.

\paragraph{Pair Classification (4 tasks).}
Duplicate detection (SprintDuplicateQuestions), paraphrase identification (TwitterURLCorpus), legal matching (LegalBenchPC), and entailment (RTE3, multilingual).
\emph{LLM:} Binary output per pair with reasoning.
\emph{Embedding:} Cosine similarity with threshold sweep (MTEB convention; best threshold chosen post-hoc on the test set).
\emph{Metric:} Average precision and accuracy.

\paragraph{Retrieval (6 tasks).}
A domain-diverse suite: legal (AILAStatutes; LegalBench consumer-contracts QA \citep{guha2023legalbench}), finance (HC3-Finance \citep{guo2023hc3}), public-health QA, French QA (FQuAD \citep{dhoffschmidt2020fquad}), and Danish social media (TwitterHjerne \citep{holm2024danoliterate}).
\emph{LLM:} Corpus-in-context \citep{lee2024longcontextlanguagemodelssubsume}: the full corpus is placed in the prompt with sequential identifiers; prompt caching \citep{gim2024promptcache} amortises the corpus prefix across queries.
Corpora are deliberately small (82--415 documents) so that models with shorter context windows can be evaluated on identical data; this keeps the whole corpus readable in one prompt, which production-scale retrieval does not permit (see \S\ref{sec:limitations}).
\emph{Embedding:} Cosine similarity ranking.
\emph{Metric:} Recall@1 (nDCG@$k$ results in released data files).

\subsection{Cost Methodology}
\label{sec:cost}

\paragraph{Embedding costs.}
We measure maximum-throughput token processing on an H100 80GB at sequence length~512 (largest fitting batch) and compute cost as $r_{\text{GPU}} / T$, where $r_{\text{GPU}} = \$2.49$/hr (Lambda Labs H100 spot, March 2026) and $T$ is tokens per hour.
Costs range from \$0.001 (mE5-small) to \$0.22 (F2LLM-v2-14B); full per-model throughput and cost data appear in Table~\ref{tab:embedding_throughput} (Appendix~\ref{app:emb_throughput}).

\paragraph{LLM costs.}
Token usage is extracted from each provider's \texttt{usage\_stats}. We bill every non-input (generated) token at the output rate:
\begin{equation}
\text{Cost} = (\text{input} - \text{cached}) \cdot r_{\text{in}} + \text{cached} \cdot r_{\text{cache}} + (\text{total} - \text{input}) \cdot r_{\text{out}}
\label{eq:cost}
\end{equation}
where $r_{\text{cache}} = r_{\text{in}} / 10$. Reasoning (``thinking'') tokens are part of the generated total and are billed at the output rate, matching provider billing; providers report them differently, and we attribute each model's reasoning share from its usage statistics (Appendix~\ref{app:cost_detail}).
Gemini models are priced at Gemini API rates and open-weight models at OpenRouter public rates (March/June 2026).
Costs range from \$3.16 (DeepSeek-V4-Flash) to \$154.14 (Gemini~3.1~Pro); the complete token-level breakdown is in Table~\ref{tab:llm_tokens} (Appendix~\ref{app:tokens_table}).

\subsection{Throughput Methodology}
\label{sec:throughput}

We serve both paradigms on the same GPU, giving the throughput comparison common hardware and removing API rate limits.
This differs from our cost accounting (\S\ref{sec:cost}), which uses API rates for LLMs and GPU-rental rates for embeddings.
We serve the two open-weight LLMs that fit on a single H100 (Qwen3.6-27B dense, Qwen3.6-35B-A3B MoE) under vLLM, and run embedding models batched on the same GPU, measuring both in tokens per second (configuration in Appendix~\ref{app:cost_detail}).
Gemini is API-only, and larger open models such as DeepSeek-R1 need multiple GPUs.
This end-to-end comparison uses each paradigm's standard inference procedure: encoder batching for embeddings and autoregressive decoding for LLMs (Appendix~\ref{app:extended_limitations}).
Per-model embedding throughput is reported in Table~\ref{tab:embedding_throughput} (Appendix~\ref{app:emb_throughput}); the comparison appears in Figure~\ref{fig:throughput} (Appendix~\ref{app:figures}).

\section{Results}
\label{sec:results}

\subsection{Overall Performance}
\label{sec:overall}

Across all 36 models, the best LLM and the best embedding model are effectively tied.
Gemini~3.1~Pro leads with a mean score of 77.6 on \mteblm{}, followed closely by Octen-8B (77.2) and Qwen3-E-8B (77.0; full ranking in Appendix~\ref{app:models}).
Gemini~3~Flash scores 75.2, and Flash-Lite scores 64.5. The reasoning-capable Gemini models occupy the top LLM ranks.

\subsection{Task-Category Analysis}
\label{sec:categories}

The aggregate tie hides large differences between task categories.
Table~\ref{tab:main_results} reports per-category scores for the top models, and full per-category rankings for all 36 models appear in Appendix~\ref{app:category_rankings}.
Figure~\ref{fig:taskrange} (Appendix~\ref{app:figures}) shows this at the task level: some embedding model matches the best LLM on 7 of 8 classification and 7 of 10 STS tasks, but on only 1 of 6 retrieval tasks.
Retrieval is where the paradigms separate; elsewhere the choice is about cost, not quality.

\begin{table*}[t]
\centering
\small
\setlength{\tabcolsep}{4pt}
\resizebox{\textwidth}{!}{%
\begin{tabular}{llcccccccr}
\toprule
& \textbf{Model} & \textbf{Params} & \textbf{Cls (8)} & \textbf{Clust (9)} & \textbf{STS (10)} & \textbf{PairCls (4)} & \textbf{Retr (6)} & \textbf{Overall (37)} & \textbf{Cost} \\
\midrule
\multirow{10}{*}{\rotatebox{90}{\small LLM}}
& Gemini 3.1 Pro & -- & 85.2 & 66.6 & 88.5 & 83.2 & \textbf{64.5} & \textbf{77.6} & \$154 \\
& Gemini 3 Flash & -- & 84.1 & 65.7 & 87.6 & 86.3 & 52.3 & 75.2 & \$56 \\
& Qwen3.6-27B & -- & 84.8 & 53.6 & 84.9 & 83.6 & 62.4 & 73.9 & \$103 \\
& Qwen3.6-35B-A3B & -- & 83.8 & 55.3 & 84.0 & 82.9 & 60.4 & 73.3 & \$34 \\
& Kimi-K2.6 & -- & 84.5 & 55.4 & 83.9 & 78.0 & 56.7 & 71.7 & \$111 \\
& DeepSeek-V4-Flash & -- & 81.9 & 43.7 & 82.2 & 81.6 & 53.3 & 68.5 & \$3 \\
& MiniMax-M2.7 & -- & 80.9 & 48.6 & 81.0 & 82.8 & 48.0 & 68.2 & \$25 \\
& GLM-4.7 & -- & 84.2 & 36.5 & 83.5 & 79.7 & 54.4 & 67.7 & \$63 \\
& Gemini 3.1 Flash Lite & -- & 82.9 & 21.7 & 85.1 & 83.7 & 49.0 & 64.5 & \$7 \\
& DeepSeek-R1 & -- & 82.5 & 31.3 & 83.7 & 79.5 & 44.7 & 64.3 & \$57 \\
\midrule
\multirow{10}{*}{\rotatebox{90}{\small Embedding}}
& Octen-8B & 7.6B & 90.1 & 65.1 & 88.7 & 86.1 & 56.0 & 77.2 & \$0.11 \\
& Qwen3-E-8B & 7.6B & 90.1 & 65.9 & 88.5 & 86.5 & 54.2 & 77.0 & \$0.11 \\
& Qwen3-E-4B & 4.0B & 89.3 & 64.6 & \textbf{88.8} & 86.5 & 50.3 & 75.9 & \$0.07 \\
& Nemotron-8B & 7.5B & 84.8 & 64.1 & 86.2 & 86.7 & 54.7 & 75.3 & \$0.11 \\
& KaLM-12B & 11.8B & 88.8 & 63.4 & 84.6 & \textbf{87.1} & 49.9 & 74.8 & \$0.16 \\
& Jina-v5-S & 596M & 90.4 & 61.5 & 86.9 & 85.1 & 48.0 & 74.4 & \$0.03 \\
& SFR-2 & 7.1B & \textbf{90.8} & \textbf{66.7} & 77.9 & 85.3 & 48.8 & 73.9 & \$0.14 \\
& Jina-v5-Nano & 212M & 89.6 & 60.4 & 87.0 & 85.0 & 47.4 & 73.9 & \$0.01 \\
& F2LLM-14B & 14.0B & 77.7 & 66.5 & 84.1 & 86.3 & 52.1 & 73.3 & \$0.22 \\
& GTE-Qwen2-7B & 7.1B & 86.7 & 65.2 & 81.6 & 86.0 & 45.8 & 73.1 & \$0.09 \\
\bottomrule
\end{tabular}}
\caption{\textbf{Model overview and per-category results.}
All ten LLMs and the ten highest-scoring embedding models. Scores are category means on a 0--100 scale; Overall is their mean.
Cost is one \mteblm{} pass using API rates for LLMs and H100 throughput at \$2.49/hr for embeddings (\S\ref{sec:cost}). Bold = best shown; full results are in Appendix~\ref{app:models}.}
\label{tab:main_results}
\end{table*}

\paragraph{Retrieval ($+$8.5): LLMs lead.}
Pro leads the best embedding on retrieval (64.5 vs.\ 56.0) and wins five of the six retrieval tasks, losing only legal statute retrieval (Appendix~\ref{app:retrieval_detail}).

\paragraph{Classification ($-$5.6): embeddings lead.}
SFR-2 \citep{meng2024sfremb2} (90.8) outscores Pro (85.2) by a wide margin.
The gap widens on fine-grained tasks (Banking77: 77 classes; MassiveIntent: 60 intents), a consequence of the kNN-vs-zero-shot setup we adopt for deployment realism (\S\ref{sec:tasks}); see \S\ref{sec:discussion} for the structural argument.

\paragraph{Clustering, STS, and pair classification: statistical ties.}
On clustering the best embedding (SFR-2, 66.7) edges Pro (66.6); on STS Qwen3-E-4B \citep{qwen2025qwen3embedding} (88.8) edges Pro (88.5); on pair classification KaLM-12B \citep{hu2025kalmv2} (87.1) leads Pro (83.2).
All three rest on geometric proximity or labelled-reference matching, which cosine similarity over learned embeddings captures directly.

\paragraph{Statistical significance.}
A paired bootstrap test (10{,}000 resamples) places the overall difference between Pro and Octen-8B within statistical noise ($\Delta$\,=\,$+$0.3, $p$\,=\,0.85, 95\%~CI\,=\,[$-$2.4, $+$3.1]).
Category-level tests favour LLMs on retrieval and embeddings on classification; clustering, STS, and pair classification are statistically even (Appendix~\ref{app:significance}, Table~\ref{tab:significance}).

\begin{figure*}[t]
\centering
\includegraphics[width=\textwidth]{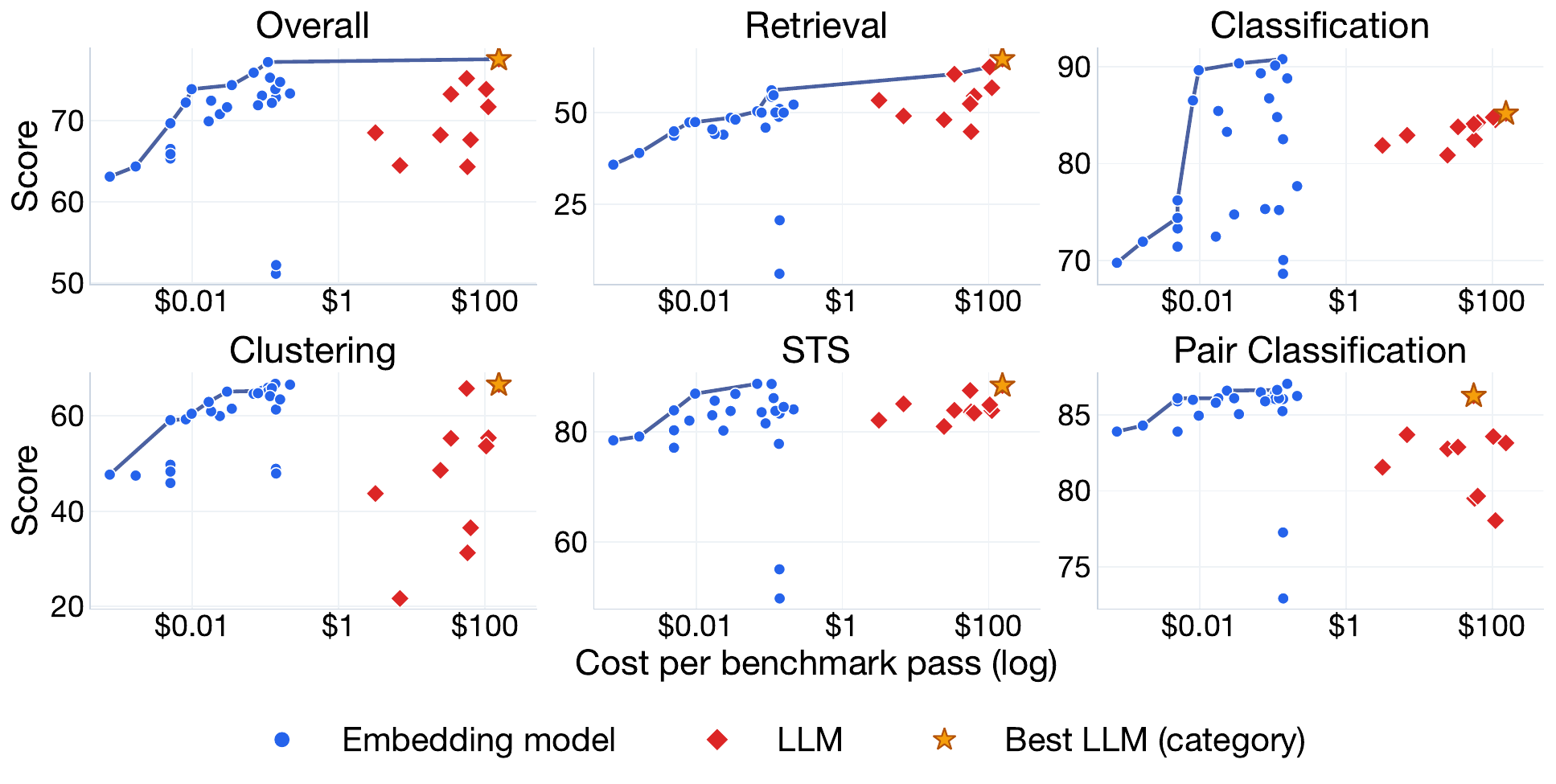}
\caption{\textbf{Cost vs.\ performance by task category.}
Each panel plots score (0--100) against cost per benchmark pass (log scale) for all 36 models; the line is the Pareto frontier over all models and the star marks the best LLM in that category.
Pro extends the retrieval frontier; embedding models define the frontiers for classification, clustering, STS, and pair classification.}
\label{fig:paretocat}
\end{figure*}

The category-level cost--performance frontiers show the same division (Figure~\ref{fig:paretocat}): Pro extends the retrieval frontier, while embeddings define the frontiers for classification, clustering, STS, and pair classification.

\subsection{Retrieve-then-Rerank}
\label{sec:rerank}

The corpus-in-context comparison above pits single-vector embeddings against LLMs directly; production systems often insert a middle stage: a cross-encoder or LLM reranker over a first-stage shortlist.
We evaluate this pipeline on the standard IR benchmarks BEIR (semantic) and BRIGHT (reasoning-heavy), crossing four first-stage retrievers with cross-encoder and LLM listwise rerankers (Figure~\ref{fig:rerank}; the full matrix is in Appendix~\ref{app:rerank_matrix}, Table~\ref{tab:reranker_matrix}).
The result follows the same task-dependent pattern: on reasoning-heavy BRIGHT, an LLM listwise reranker improves a strong embedding first stage from 22.3 to 35.1 nDCG@10.
On semantic BEIR, the strong embedding alone scores 63.1, ahead of the best reranked configuration at 60.3.
Cost scales with shortlist size: reranking a top-100 shortlist costs \$10--30 per benchmark, against \$154 for reading the full corpus in context.
The MoE Qwen3.6-35B-A3B is cheaper still, reaching 33.6 nDCG@10 on BRIGHT for \$10 where Qwen3.6-27B reaches 35.1 for \$30.

\begin{figure}[t]
\centering
\includegraphics[width=\linewidth]{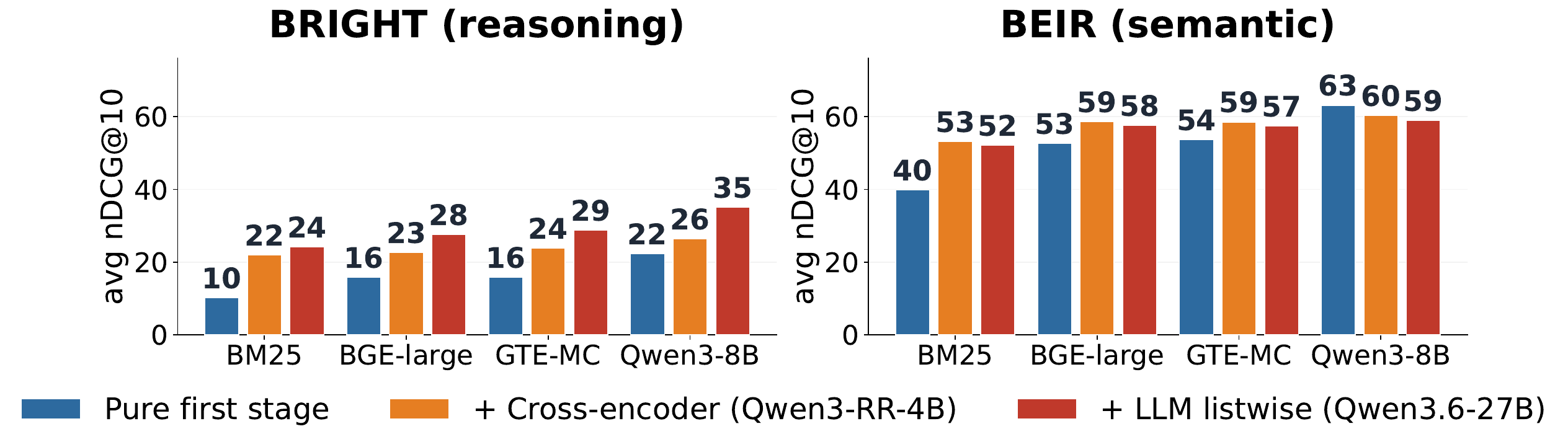}
\caption{\textbf{Retrieve-then-rerank.}
LLM listwise reranking improves every first stage on BRIGHT. On BEIR, a strong embedding first stage outperforms all reranked configurations.}
\label{fig:rerank}
\end{figure}

\subsection{Cost and Throughput}
\label{sec:cost_results}

\paragraph{Embeddings dominate the Pareto frontier.}
The Pareto frontier contains the embedding models and Gemini~3.1~Pro, which costs 1{,}431$\times$ as much as the best embedding of comparable quality (Figure~\ref{fig:pareto}).
Pro's marginal 0.4-point aggregate edge over Octen-8B comes with a benchmark cost of \$154.14 versus \$0.11.

\begin{figure}[t]
\centering
\includegraphics[width=0.95\textwidth]{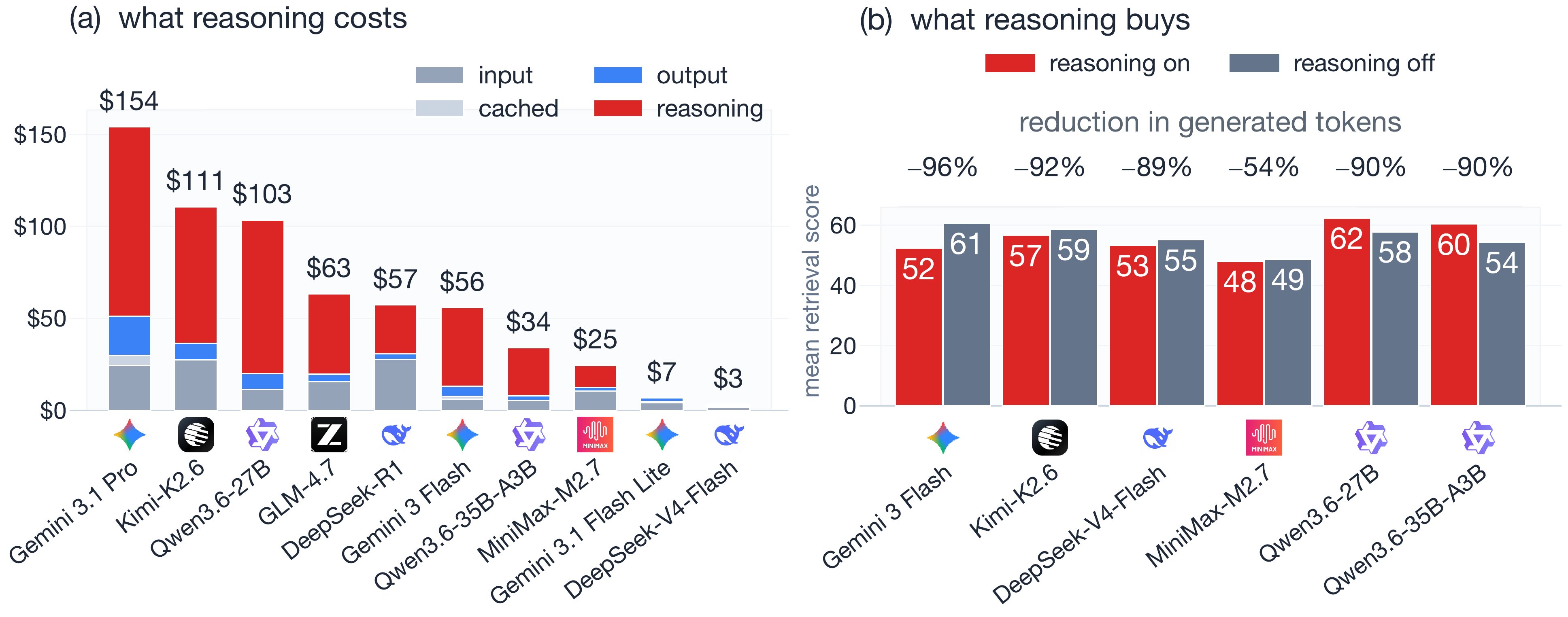}
\caption{\textbf{The thinking-token tax: what reasoning costs, and what it buys.}
\emph{(a)} API cost per benchmark pass by token type for all ten LLMs.
\emph{(b)} Mean retrieval score with default vs.\ disabled reasoning for six models from five families; labels show the reduction in generated tokens.
Four models preserve or improve retrieval with 54--96\% fewer generated tokens; the two Qwen models lose ground.}
\label{fig:cost_breakdown}
\end{figure}

\paragraph{The thinking-token tax.}
The structural driver of LLM cost is internal reasoning (Equation~\ref{eq:cost}; Figure~\ref{fig:cost_breakdown}).
Reasoning tokens account for 28--81\% of inference cost across the reasoning models. Flash-Lite provides a non-reasoning reference point and ranks below most embeddings. This spending buys little on most tasks: lower reasoning budgets preserve or improve retrieval for four of six models (\S\ref{sec:ablation}).

\paragraph{Cost sensitivity.}
The 1{,}431$\times$ ratio assumes H100 spot pricing (\$2.49/hr).
Under alternative hardware and pricing the ratio ranges from 338$\times$ (commercial embedding API at \$0.10/MTok) to 2{,}424$\times$ (L4 GPU at \$0.49/hr).
The order-of-magnitude gap persists across all evaluated cost assumptions (Appendix~\ref{app:cost_sensitivity}).

\subsection{Throughput Constraints}
\label{sec:throughput_results}

Served on the same H100, embedding models process tokens far faster than LLMs (Figure~\ref{fig:throughput}, Appendix~\ref{app:figures}).
The two open-weight LLMs that fit on a single H100 (Qwen3.6-27B, Qwen3.6-35B-A3B) sustain 5{,}400--5{,}900 tokens/second, whereas embedding models range from 14{,}700 tok/s (the largest, F2LLM-14B) to 4.3M tok/s (the smallest, mE5-small), a 2.5$\times$ to 736$\times$ advantage on identical hardware.
With hardware and serving stack fixed, the measured gap reflects the different inference procedures: autoregressive decoding and a single encoder pass.
At production scale (millions of documents and continuous ingestion), this throughput gap limits the practicality of LLM-based pipelines.

\subsection{Ablation: Reduced Thinking}
\label{sec:ablation}

We re-evaluate Gemini~3~Flash with \texttt{reasoning\_effort=low} and, for the open models, with reasoning disabled at the serving layer, cutting reasoning tokens by 54--96\% on the retrieval tasks (Table~\ref{tab:ablation_nocot}; cross-family results in Figure~\ref{fig:cost_breakdown}b; details in Appendix~\ref{app:ablation_detail}).
Two findings stand out.
First, reduced reasoning preserves or improves all six Flash retrieval scores and four of six cross-family averages.
Second, classification changes by less than one point on all tested tasks ($|\Delta| < 1.0$).
The retrieval advantage largely persists under reduced reasoning budgets, indicating a limited role for additional test-time reasoning.

\subsection{Ablation: Few-Shot Classification}
\label{sec:fewshot}

To test whether labelled examples can close the classification gap, we evaluate Flash with 5 in-context examples per task (Table~\ref{tab:ablation_fewshot}; details in Appendix~\ref{app:ablation_detail}).
Five-shot results are similar to or worse than the zero-shot scores on the small-label tasks. Note that the embedding kNN classifier uses the full labelled train set. The LLM prompt only gets five examples here, yet given these results, it seems unlikely more would help.

\section{Discussion}
\label{sec:discussion}

In aggregate, the two paradigms tie on maximum performance. The LLM advantage is narrow and specific: reasoning-heavy retrieval, where bi-encoders score systematically low \citep{su2024bright}.
Pro leads on 5/6 retrieval datasets, showing a consistent advantage across the suite.
Capability profiles show this split: LLMs lead on retrieval while embeddings provide balanced, often superior, coverage across the other four categories (Appendix~\ref{app:figures}, Figure~\ref{fig:radar}).

\begin{figure}[t]
\centering
\includegraphics[width=0.95\textwidth]{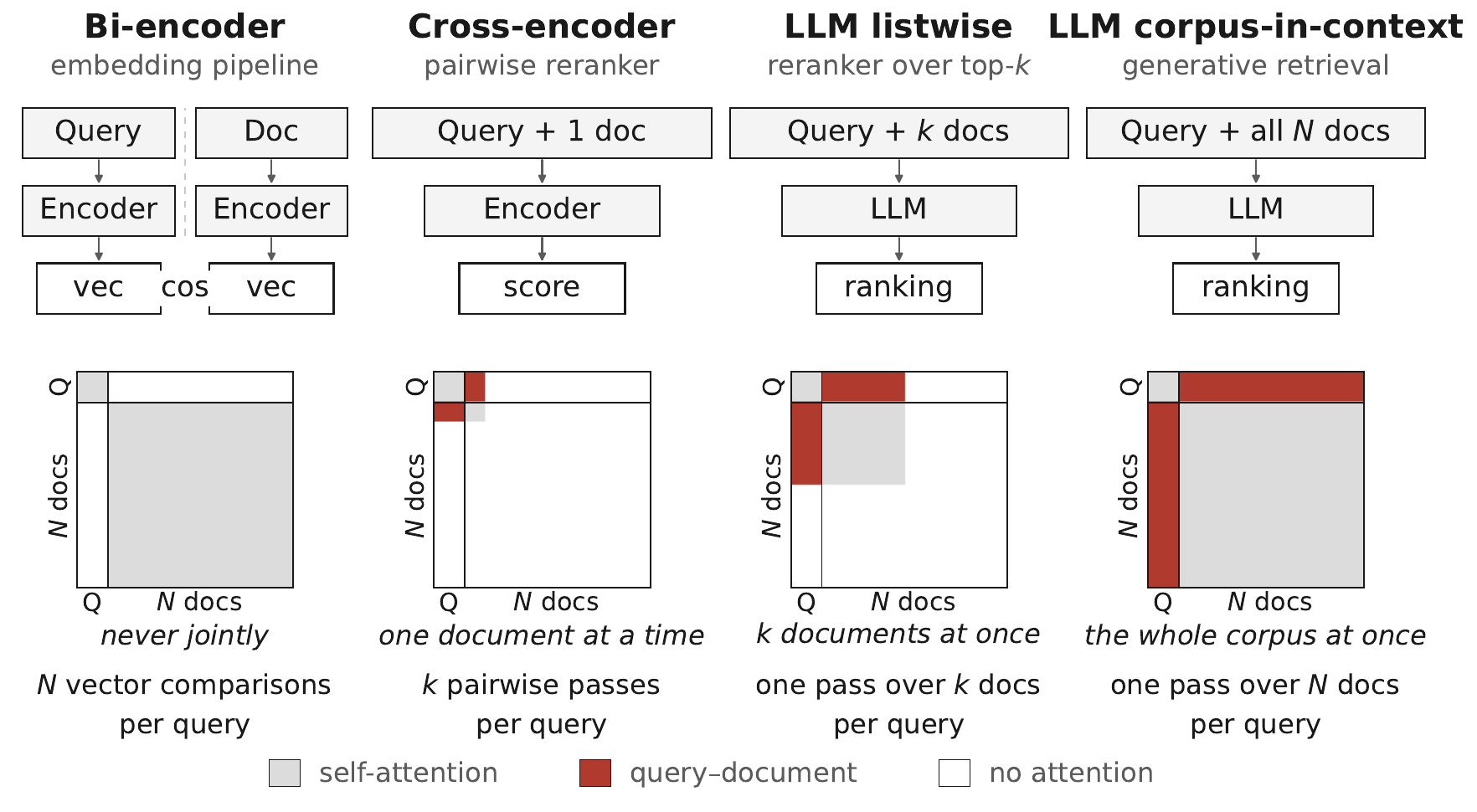}
\caption{\textbf{How many documents each architecture reads jointly with the query.}
\emph{Top:} the pipeline. \emph{Bottom:} its attention mask, drawn over the same $N$ documents at the same scale in every panel so the four are directly comparable.
Each row is a token and each column a token it may read.
The red region is what one forward pass reads jointly with the query; it cannot be computed before the query arrives, so cost grows with it.}
\label{fig:architecture}
\end{figure}

Viewed through the lens of information retrieval, our results reproduce the classic bi-encoder versus cross-encoder tradeoff at LLM scale~\citep{reimers2019sentencebert, muennighoff2022sgptgptsentenceembeddings}.
One design choice separates the four architectures we evaluate (Figure~\ref{fig:architecture}): how many documents the model is allowed to read jointly with the query.
Embedding pipelines encode queries and documents independently: each document is processed once, offline, and each query requires only vector comparisons.
A cross-encoder reranker reads one document at a time, an LLM listwise reranker reads the top-$k$ shortlist at once, and an LLM in corpus-in-context mode is the extreme case, reading the entire corpus in a single forward pass.
Cost follows that ordering because query-specific computation repeats for each query: reranking scales with the shortlist $k$, and corpus-in-context processing scales with the corpus $N$.
Quality follows the same ordering, which is why reranking is a more economical way to add reasoning than placing the full corpus in context (\S\ref{sec:rerank}).

Our retrieve-then-rerank experiment (\S\ref{sec:rerank}) shows the same pattern in a production-style setting: a reranker helps most on reasoning-heavy retrieval and little on semantic retrieval.
The same-hardware results attribute much of the quality--cost gap to the underlying inference procedures.

The classification pipelines receive different supervision. Embedding classifiers use kNN over the full labelled training distribution; the LLM receives label names, task descriptions, and up to five examples. Fine-grained tasks make this difference especially visible (\S\ref{sec:fewshot}).
Leading embedding models are also contrastively trained on data overlapping these domains and label spaces, so part of the gap reflects in-domain fit rather than a general representational advantage.
The comparison is therefore between deployment pipelines as practitioners meet them, not intrinsic ceilings: a classification-post-trained LLM would likely narrow the gap, and none was available among the frontier models we evaluate.

The thinking-token tax reveals a mismatch between default reasoning budgets and these tasks.
Reasoning improves the overall Flash score, while lower reasoning budgets preserve retrieval quality for most evaluated models.
Where direct reading comprehension suffices, the model reconsiders straightforward relevance judgements, making the default reasoning budget wasteful for most of these benchmark tasks \citep{lu2025cothurts}.
For practitioners, reducing reasoning effort preserves retrieval quality for most evaluated models while removing the largest component of LLM cost.

\paragraph{Deployment implications.}
These findings motivate a hybrid architecture: embedding models provide high-throughput candidate retrieval, followed by LLM reasoning over a shortlist (Figure~\ref{fig:throughput}).
This division of labour supports retrieve-then-reason pipelines \citep{lewis2020rag, gao2024retrievalaugmentedgenerationlargelanguage}; our reranker results (\S\ref{sec:rerank}) show where they help.
Across the four non-retrieval categories, small-to-medium embedding models closely match the best LLM at a fraction of the cost.
Hardware improvements will reduce the absolute cost of both paradigms. Their different inference procedures preserve an embedding throughput advantage for similarity, classification, and clustering workloads (\S\ref{sec:throughput}).
Accuracy-only leaderboards obscure large cost differences between similarly scored systems. We recommend reporting Pareto frontiers and significance tests alongside accuracy \citep{card2020reporting, dehghani2021benchmarklottery}.

\paragraph{Limitations.}
\label{sec:limitations}
Our ten LLMs from six families are a snapshot of a fast-moving frontier; because \mteblm{} uses the MTEB framework, future models can be evaluated with the same pipeline.
The corpus-in-context protocol places corpora of 82--415 documents entirely in the prompt, which is only feasible at small scale: production corpora require an indexed first stage, so the cost gap we report is a lower bound.
We complement this setting with a bi-encoder, cross-encoder reranker, and LLM listwise reranker comparison on BEIR and BRIGHT (\S\ref{sec:rerank}).
Embeddings also benefit from post-hoc threshold optimisation in pair classification (MTEB convention) with no LLM analogue.
Appendix~\ref{app:extended_limitations} gives the full set.

\section{Conclusion}
\label{sec:conclusion}

We present \mteblm{}, the first cost-aware comparison of LLM and embedding deployment pipelines across five MTEB task categories, evaluating ten LLMs from six families and 26 embedding models on 37 tasks.
The two paradigms match in aggregate but differ by task: LLMs lead on reasoning-heavy retrieval, embeddings lead on classification, and the rest are tied. Comparable LLM quality comes with substantially higher cost and lower throughput.

Embedding pipelines therefore remain the cost-efficient default, with reasoning-intensive retrieval the clearest case for an LLM.
Our retrieve-then-rerank results support a hybrid strategy: embeddings for candidate retrieval, LLMs for reasoning over the shortlist.

Pareto frontiers add deployment cost and throughput to accuracy-based rankings, and we encourage their wider use.
\mteblm{} is released with MTEB-compatible code and datasets, so the evaluation can expand as new models appear.

\section*{Acknowledgments}

We are extremely thankful to Laude Institute for supporting this work.

\section*{Reproducibility Statement}

Code, raw results, and analysis scripts are released at \url{https://github.com/embeddings-benchmark/embedders-dilemma} in an implementation built on the MTEB framework.
The LLM-specific datasets are hosted on Hugging Face under \texttt{mteb/llm-eval-*}, and each task pins its dataset to an exact revision.
LLM prompt templates, schema validation logic, and token-usage extraction scripts are included in the release, and every figure and table in the paper regenerates from the released result files.
The provided scripts reproduce the embedding throughput benchmarks on an NVIDIA H100 GPU.
Detailed experimental configurations are described in \S\ref{sec:setup} and Appendix~\ref{app:reproducibility}.

\section*{Ethics Statement}

This work compares publicly available models on standard benchmarks and does not involve human subjects or private data.
All evaluated datasets are publicly hosted and have been previously released for research purposes.
Cost and throughput reporting may influence deployment decisions; we encourage practitioners to consider the environmental impact of LLM-based pipelines, including the carbon footprint of chain-of-thought inference, alongside the economic factors discussed in this paper.
Reducing reasoning budgets can preserve performance on some tasks while lowering compute use.
LLM assistance was used during the preparation of this paper for writing polish, phrasing refinement, and code formatting (e.g., matplotlib styling); no LLM was used to originate research ideas, generate evaluation data, produce plots, or evaluate model outputs. All scientific content, experimental design, and conclusions are the authors' own.

\bibliographystyle{colm2026_conference}
\bibliography{colm2026_conference}

\appendix


\section{Statistical Significance Testing}
\label{app:significance}

\paragraph{Bootstrap procedure.}
We implement a paired bootstrap test \citep{efron1979bootstrap} comparing Gemini~3.1~Pro against the best embedding model in each comparison.
The resampling unit is the task level: each bootstrap replicate draws \mteblm{} tasks with replacement from the full \mteblm{} task set, and the score difference is computed as the mean-of-means difference across the resampled task set.
Task-level resampling accommodates the different metrics used across categories (accuracy, Spearman $\rho$, V-measure, average precision, and Recall@1). Equal task weighting matches the macro-average score.

We report unadjusted $p$-values for five planned category-level comparisons. Under a Bonferroni threshold of $\alpha$\,=\,0.01, classification remains significant and retrieval falls above the threshold.

Table~\ref{tab:significance} reports the full results.
The overall comparison (Pro vs.\ Octen-8B across \mteblm) yields $\Delta$\,=\,$+$0.3, 95\%~CI\,=\,[$-$2.4, +3.1], $p$\,=\,0.85, a statistical tie.
Category-level tests identify two significant differences:
\begin{itemize}
    \item \textbf{Classification:} embeddings win ($\Delta$\,=\,$-$5.6, CI\,=\,[$-$9.2, $-$2.4], $p$\,$<$\,0.01)
    \item \textbf{Retrieval:} LLMs win ($\Delta$\,=\,+8.5, CI\,=\,[+0.2, +16.8], $p$\,$<$\,0.05)
\end{itemize}
STS ($p$\,=\,0.74), clustering ($p$\,=\,0.97), and pair classification ($p$\,=\,0.49) show no significant paradigm difference.

\begin{table}[ht]
\centering
\small
\begin{tabular}{lcccc}
\toprule
\textbf{Comparison} & \textbf{$\Delta$} & \textbf{95\% CI} & \textbf{$p$} & \textbf{Sig.} \\
\midrule
\multicolumn{5}{l}{\emph{Overall (Gemini 3.1 Pro vs.\ Octen-8B, \mteblm)}} \\
\quad All tasks & +0.3 & [$-$2.4, +3.1] & 0.85 & No \\
\midrule
\multicolumn{5}{l}{\emph{Per category (Gemini 3.1 Pro vs.\ best embedding)}} \\
\quad Retrieval & +8.5 & [+0.2, +16.8] & $<$0.05 & Yes \\
\quad Clustering & $-$0.2 & [$-$5.6, +5.1] & 0.96 & No \\
\quad STS & $-$0.3 & [$-$2.2, +1.8] & 0.75 & No \\
\quad Pair Classification & $-$3.9 & [$-$13.6, +5.9] & 0.50 & No \\
\quad Classification & $-$5.6 & [$-$9.2, $-$2.4] & $<$0.01 & Yes \\
\bottomrule
\end{tabular}
\caption{\textbf{Statistical significance.} Paired bootstrap test (10{,}000 resamples, seed 42).
$\Delta$ = Gemini 3.1 Pro $-$ best embedding; significance at $\alpha$\,=\,0.05. Pair classification uses Pro for consistency, although Flash scores higher.}
\label{tab:significance}
\end{table}

\section{Evaluated Models and Tasks}
\label{app:models_detail}

This section lists the evaluated models, category- and task-level results, and benchmark tasks and metrics.

\subsection{Model Overview}

Table~\ref{tab:full_models} lists all 36 evaluated models — 10 LLMs and 26 embedding models — with parameter counts, mean \mteblm{} scores, total benchmark costs, and paper references.
Models range from mE5-small (118M parameters, \$0.001) to Gemini~3.1~Pro (\$154.14).
Models without a dedicated paper are marked ``--''.

\begin{table}[ht]
\centering
\small
\begin{tabular}{lcrrl}
\toprule
\textbf{Model} & \textbf{Params} & \textbf{Score} & \textbf{Cost} & \textbf{Reference} \\
\midrule
\multicolumn{5}{l}{\emph{LLM Models}} \\
Gemini 3.1 Pro & -- & \textbf{77.6} & \$154.14 & -- \\
Gemini 3 Flash & -- & 75.2 & \$55.87 & -- \\
Qwen3.6-27B & -- & 73.9 & \$103.43 & -- \\
Qwen3.6-35B-A3B & -- & 73.3 & \$34.06 & -- \\
Kimi-K2.6 & -- & 71.7 & \$110.70 & -- \\
DeepSeek-V4-Flash & -- & 68.5 & \$3.16 & -- \\
MiniMax-M2.7 & -- & 68.2 & \$24.52 & -- \\
GLM-4.7 & -- & 67.7 & \$63.31 & -- \\
Gemini 3.1 Flash Lite & -- & 64.5 & \$6.85 & -- \\
DeepSeek-R1 & -- & 64.3 & \$57.38 & -- \\
\midrule
\multicolumn{5}{l}{\emph{Embedding Models (ranked by score)}} \\
Octen-8B & 7.6B & 77.2 & \$0.108 & -- \\
Qwen3-E-8B & 7.6B & 77.0 & \$0.108 & \citealt{qwen2025qwen3embedding} \\
Qwen3-E-4B & 4.0B & 75.9 & \$0.069 & \citealt{qwen2025qwen3embedding} \\
Nemotron-8B & 7.5B & 75.3 & \$0.115 & \citealt{lee2024nvembedimprovedtechniques} \\
KaLM-12B & 11.8B & 74.8 & \$0.158 & \citealt{hu2025kalmv2} \\
Jina-v5-S & 596M & 74.4 & \$0.034 & \citealt{akram2026jinaembeddingsv5} \\
SFR-2 & 7.1B & 73.9 & \$0.136 & \citealt{meng2024sfremb2} \\
Jina-v5-Nano & 212M & 73.9 & \$0.010 & \citealt{akram2026jinaembeddingsv5} \\
F2LLM-14B & 14.0B & 73.3 & \$0.215 & \citealt{zhang2026f2llmv2} \\
GTE-Qwen2-7B & 7.1B & 73.1 & \$0.089 & \citealt{li2023generaltextembeddingsmultistage} \\
Linq-Mistral & 7.1B & 72.8 & \$0.138 & \citealt{choi2024linq} \\
Qwen3-E-0.6B & 596M & 72.5 & \$0.018 & \citealt{qwen2025qwen3embedding} \\
EmbGemma-300M & 308M & 72.2 & \$0.008 & \citealt{lee2025geminiembedding} \\
F2LLM-8B & 7.6B & 72.2 & \$0.122 & \citealt{zhang2026f2llmv2} \\
F2LLM-4B & 4.0B & 71.9 & \$0.079 & \citealt{zhang2026f2llmv2} \\
F2LLM-1.7B & 1.7B & 71.7 & \$0.030 & \citealt{zhang2026f2llmv2} \\
GTE-Qwen2-1.5B & 1.5B & 70.8 & \$0.024 & \citealt{li2023generaltextembeddingsmultistage} \\
F2LLM-0.6B & 596M & 69.9 & \$0.017 & \citealt{zhang2026f2llmv2} \\
mE5-L-Inst & 560M & 69.7 & \$0.005 & \citealt{wang2024improvingtextembeddingslarge} \\
BGE-M3 & 568M & 66.5 & \$0.005 & \citealt{chen2025m3embeddingmultilingualitymultifunctionalitymultigranularity} \\
mE5-L & 560M & 65.9 & \$0.005 & \citealt{wang2024textembeddingsweaklysupervisedcontrastive} \\
Arctic-L-v2 & 568M & 65.3 & \$0.005 & \citealt{yu2024arctic} \\
mE5-B & 278M & 64.4 & \$0.002 & \citealt{wang2024textembeddingsweaklysupervisedcontrastive} \\
mE5-S & 118M & 63.1 & \$0.001 & \citealt{wang2024textembeddingsweaklysupervisedcontrastive} \\
E5-Mistral-7B & 7.1B & 52.2 & \$0.139 & \citealt{wang2024improvingtextembeddingslarge} \\
GritLM-7B & 7.2B & 51.2 & \$0.138 & \citealt{muennighoff2025generativerepresentationalinstructiontuning} \\
\bottomrule
\end{tabular}
\caption{\textbf{Complete model listing} with mean (macro) scores across 37 \mteblm{}
tasks, total benchmark costs, and references. LLM costs reflect actual API usage; embedding costs
from H100 throughput benchmarking (\$2.49/hr).}
\label{tab:full_models}
\end{table}

\subsection{Full Per-Category Rankings}
\label{app:models}

Table~\ref{tab:category_scores} compares Pro with the best embedding in each category. Table~\ref{tab:full_results_by_category} gives per-category scores for all 36 models.
Pro ranks first overall largely because of retrieval; on classification, it ranks below ten embedding models.
The reasoning-capable Gemini models rank far above the non-reasoning Flash-Lite, which places 31st of 36 overall.

\begin{table}[ht]
\centering
\small
\begin{tabular}{lccccc}
\toprule
\textbf{Category} & \textbf{Gemini 3.1 Pro} & \textbf{Gemini 3 Flash} & \textbf{Best Emb.} & \textbf{Best Model} & \textbf{$\Delta$} \\
\midrule
Retrieval (6) & \textbf{64.5} & 52.3 & 56.0 & Octen-8B & +8.5 \\
Clustering (9) & 66.6 & 65.7 & \textbf{66.7} & SFR-2 & $-$0.2 \\
STS (10) & 88.5 & 87.6 & \textbf{88.8} & Qwen3-E-4B & $-$0.3 \\
PairCls (4) & 83.2 & 86.3 & \textbf{87.1} & KaLM-12B & $-$3.9 \\
Classification (8) & 85.2 & 84.1 & \textbf{90.8} & SFR-2 & $-$5.6 \\
\bottomrule
\end{tabular}
\caption{\textbf{Category-level performance} (best LLM vs.\ best embedding per category).
$\Delta$ = best LLM $-$ best embedding; bold = winner; task counts in parentheses.}
\label{tab:category_scores}
\end{table}

\begin{table*}[ht]
\centering
\small
\setlength{\tabcolsep}{3pt}
\begin{tabular}{rlccccccc}
\toprule
\textbf{Rank} & \textbf{Model} & \textbf{Type} & \textbf{Cls (8)} & \textbf{Clust (9)} & \textbf{STS (10)} & \textbf{PairCls (4)} & \textbf{Retr (6)} & \textbf{Overall} \\
\midrule
1 & Gemini 3.1 Pro & LLM & 85.2 & 66.6 & 88.5 & 83.2 & \textbf{64.5} & \textbf{77.6} \\
2 & Octen-8B & Emb & 90.1 & 65.1 & 88.7 & 86.1 & 56.0 & 77.2 \\
3 & Qwen3-E-8B & Emb & 90.1 & 65.9 & 88.5 & 86.5 & 54.2 & 77.0 \\
4 & Qwen3-E-4B & Emb & 89.3 & 64.6 & \textbf{88.8} & 86.5 & 50.3 & 75.9 \\
5 & Nemotron-8B & Emb & 84.8 & 64.1 & 86.2 & 86.7 & 54.7 & 75.3 \\
6 & Gemini 3 Flash & LLM & 84.1 & 65.7 & 87.6 & 86.3 & 52.3 & 75.2 \\
7 & KaLM-12B & Emb & 88.8 & 63.4 & 84.6 & \textbf{87.1} & 49.9 & 74.8 \\
8 & Jina-v5-S & Emb & 90.4 & 61.5 & 86.9 & 85.1 & 48.0 & 74.4 \\
9 & SFR-2 & Emb & \textbf{90.8} & \textbf{66.7} & 77.9 & 85.3 & 48.8 & 73.9 \\
10 & Jina-v5-Nano & Emb & 89.6 & 60.4 & 87.0 & 85.0 & 47.4 & 73.9 \\
11 & Qwen3.6-27B & LLM & 84.8 & 53.6 & 84.9 & 83.6 & 62.4 & 73.9 \\
12 & F2LLM-14B & Emb & 77.7 & 66.5 & 84.1 & 86.3 & 52.1 & 73.3 \\
13 & Qwen3.6-35B-A3B & LLM & 83.8 & 55.3 & 84.0 & 82.9 & 60.4 & 73.3 \\
14 & GTE-Qwen2-7B & Emb & 86.7 & 65.2 & 81.6 & 86.0 & 45.8 & 73.1 \\
15 & Linq-Mistral & Emb & 82.5 & 61.3 & 83.4 & 86.1 & 51.0 & 72.8 \\
16 & Qwen3-E-0.6B & Emb & 85.4 & 60.9 & 85.7 & 86.1 & 44.1 & 72.5 \\
17 & EmbGemma-300M & Emb & 86.5 & 59.3 & 82.1 & 86.0 & 47.3 & 72.2 \\
18 & F2LLM-8B & Emb & 75.2 & 65.8 & 83.9 & 86.1 & 50.0 & 72.2 \\
19 & F2LLM-4B & Emb & 75.3 & 64.7 & 83.6 & 85.9 & 49.9 & 71.9 \\
20 & Kimi-K2.6 & LLM & 84.5 & 55.4 & 83.9 & 78.0 & 56.7 & 71.7 \\
21 & F2LLM-1.7B & Emb & 74.7 & 65.1 & 83.8 & 86.1 & 48.5 & 71.7 \\
22 & GTE-Qwen2-1.5B & Emb & 83.3 & 60.0 & 80.3 & 86.6 & 43.9 & 70.8 \\
23 & F2LLM-0.6B & Emb & 72.5 & 62.9 & 83.1 & 85.8 & 45.4 & 69.9 \\
24 & mE5-L-Inst & Emb & 74.4 & 59.1 & 83.9 & 86.1 & 44.8 & 69.7 \\
25 & DeepSeek-V4-Flash & LLM & 81.9 & 43.7 & 82.2 & 81.6 & 53.3 & 68.5 \\
26 & MiniMax-M2.7 & LLM & 80.9 & 48.6 & 81.0 & 82.8 & 48.0 & 68.2 \\
27 & GLM-4.7 & LLM & 84.2 & 36.5 & 83.5 & 79.7 & 54.4 & 67.7 \\
28 & BGE-M3 & Emb & 76.2 & 45.9 & 80.4 & 85.9 & 44.3 & 66.5 \\
29 & mE5-L & Emb & 73.3 & 48.3 & 80.3 & 83.9 & 43.6 & 65.9 \\
30 & Arctic-L-v2 & Emb & 71.4 & 49.7 & 77.2 & 83.9 & 44.6 & 65.3 \\
31 & Gemini 3.1 Flash Lite & LLM & 82.9 & 21.7 & 85.1 & 83.7 & 49.0 & 64.5 \\
32 & mE5-B & Emb & 71.9 & 47.5 & 79.2 & 84.3 & 38.9 & 64.4 \\
33 & DeepSeek-R1 & LLM & 82.5 & 31.3 & 83.7 & 79.5 & 44.7 & 64.3 \\
34 & mE5-S & Emb & 69.8 & 47.7 & 78.5 & 83.9 & 35.8 & 63.1 \\
35 & E5-Mistral-7B & Emb & 70.0 & 47.9 & 49.8 & 72.9 & 20.6 & 52.2 \\
36 & GritLM-7B & Emb & 68.6 & 48.9 & 55.1 & 77.3 & 6.1 & 51.2 \\
\bottomrule
\end{tabular}
\caption{\textbf{Complete per-category results} for all 36 complete models,
ranked by overall (macro) score. Bold = best in column.}
\label{tab:full_results_by_category}
\end{table*}

\subsection{Per-Task Scores (Representative Models)}
\label{app:per_task}

Table~\ref{tab:per_task_scores} reports individual task scores for 5 representative models (2 LLMs, 3 embeddings) across all \mteblm{} tasks, alongside the best embedding score per task.
The scores vary substantially by task. For example, Pro is strongest on FQuAD retrieval, and SFR-2 is strongest on SprintDuplicateQuestions.
This complementarity supports the hybrid-pipeline recommendation in \S\ref{sec:discussion}.

\begin{table*}[ht]
\centering
\scriptsize
\setlength{\tabcolsep}{3pt}
\resizebox{\textwidth}{!}{%
\begin{tabular}{llcccccr}
\toprule
\textbf{Task} & \textbf{Cat.} & \textbf{Gemini 3.1 Pro} & \textbf{Gemini 3 Flash} & \textbf{Qwen3.6-27B} & \textbf{Octen-8B} & \textbf{Qwen3-E-8B} & \textbf{Best Emb.} \\
\midrule
\multicolumn{8}{l}{\emph{Classification (Accuracy)}} \\
AmazonCounterfactualClassification & Cls & 84.8 & 81.2 & 90.1 & \textbf{93.0} & 92.9 & \textbf{93.0} \\
Banking77Classification & Cls & 85.0 & 83.1 & 80.5 & 87.3 & 87.2 & \textbf{91.6} \\
ImdbClassification & Cls & \textbf{98.0} & 97.6 & 97.4 & 97.7 & 97.8 & 97.8 \\
MTOPDomainClassification & Cls & 95.5 & 96.4 & 96.9 & 98.2 & 98.1 & \textbf{99.2} \\
MassiveIntentClassification & Cls & 84.9 & 85.4 & 84.6 & 85.7 & 85.8 & \textbf{88.9} \\
MassiveScenarioClassification & Cls & 79.4 & 75.9 & 76.4 & 89.3 & 89.4 & \textbf{93.0} \\
ToxicConversationsClassification & Cls & 89.6 & 90.0 & 89.2 & 91.9 & 92.0 & \textbf{94.4} \\
TweetSentimentExtractionClassification & Cls & 64.2 & 63.0 & 63.2 & 77.9 & 77.8 & \textbf{78.9} \\
\multicolumn{8}{l}{\emph{Clustering (V-measure)}} \\
ArxivClusteringP2P & Clust & 60.1 & 60.3 & 54.5 & 61.9 & 62.5 & \textbf{64.1} \\
ArxivClusteringS2S & Clust & \textbf{62.5} & 60.1 & 41.6 & 60.5 & 61.1 & 61.5 \\
BiorxivClusteringP2PV2 & Clust & 62.5 & 60.8 & 49.9 & 61.6 & 65.3 & \textbf{77.3} \\
MedrxivClusteringP2PV2 & Clust & 52.5 & 50.6 & 48.0 & 55.9 & 55.9 & \textbf{60.7} \\
MedrxivClusteringS2SV2 & Clust & 54.0 & 50.3 & 42.7 & 53.3 & 55.2 & \textbf{57.5} \\
RedditClusteringP2P & Clust & \textbf{93.9} & 90.5 & 60.1 & 78.7 & 79.8 & 80.8 \\
StackExchangeClusteringP2PV2 & Clust & 42.9 & 49.0 & 46.1 & 59.7 & 59.5 & \textbf{60.7} \\
StackExchangeClusteringV2 & Clust & \textbf{91.8} & 90.2 & 73.6 & 85.8 & 86.3 & 86.3 \\
TwentyNewsgroupsClusteringV2 & Clust & 78.7 & \textbf{79.9} & 66.3 & 68.5 & 67.6 & 73.8 \\
\multicolumn{8}{l}{\emph{STS (Spearman $\rho$)}} \\
BIOSSES & STS & \textbf{89.6} & 88.7 & 85.0 & 84.4 & 82.0 & 88.8 \\
SICKR & STS & 86.1 & 86.8 & 76.4 & 87.9 & 88.4 & \textbf{91.1} \\
STS12 & STS & 82.6 & 79.5 & 77.1 & 87.3 & 87.3 & \textbf{87.4} \\
STS13 & STS & 91.9 & 91.5 & 90.6 & 93.9 & 94.0 & \textbf{94.6} \\
STS14 & STS & 90.5 & 89.3 & 88.0 & 90.9 & 91.0 & \textbf{91.4} \\
STS15 & STS & 94.4 & 93.9 & 91.8 & 94.3 & 94.2 & \textbf{94.5} \\
STS16 & STS & 89.0 & 88.5 & 87.1 & 92.5 & 92.5 & \textbf{92.9} \\
STS17 & STS & \textbf{94.2} & 94.1 & 91.8 & 93.6 & 93.7 & 93.7 \\
STS22v2 & STS & \textbf{73.6} & 71.7 & 72.0 & 69.5 & 68.9 & 69.5 \\
STSBenchmark & STS & 92.6 & 91.7 & 89.6 & 93.2 & 93.4 & \textbf{94.9} \\
\multicolumn{8}{l}{\emph{PairClassification (Avg. Precision)}} \\
LegalBenchPC & PairCls & 72.2 & \textbf{77.4} & 71.6 & 70.0 & 70.4 & 73.6 \\
RTE3PC & PairCls & 95.4 & \textbf{96.7} & 92.9 & 85.4 & 85.7 & 85.7 \\
SprintDuplicateQuestionsPC & PairCls & 82.2 & 83.4 & 84.6 & 99.4 & 99.6 & \textbf{99.8} \\
TwitterURLCorpusPC & PairCls & 82.8 & 87.6 & 85.2 & 89.4 & \textbf{90.2} & \textbf{90.2} \\
\multicolumn{8}{l}{\emph{Retrieval (Recall@1)}} \\
AILAStatutes & Retr & 14.5 & 5.7 & 14.2 & \textbf{23.2} & 21.8 & \textbf{23.2} \\
FQuADRetrieval & Retr & 92.0 & 88.0 & \textbf{96.0} & 69.0 & 68.0 & 72.0 \\
HC3FinanceRetrieval & Retr & \textbf{71.0} & 60.0 & 57.0 & 66.0 & 59.0 & 67.0 \\
LegalBenchConsumerContractsQA & Retr & 86.0 & 79.0 & \textbf{88.0} & 67.0 & 70.0 & 70.0 \\
PublicHealthQA & Retr & \textbf{90.0} & 49.0 & 86.0 & 81.0 & 78.0 & 85.0 \\
TwitterHjerneRetrieval & Retr & \textbf{33.4} & 32.4 & 33.0 & 29.9 & 28.6 & 31.5 \\
\bottomrule
\end{tabular}
}
\caption{\textbf{Per-task scores} for representative models across all 37 \mteblm{} tasks.
Bold = best in row (incl.\ best embedding). Metric per category in italics.}
\label{tab:per_task_scores}
\end{table*}

\subsection{Full Per-Task Scores for All Models}
\label{app:full_per_task_all}

Tables~\ref{tab:full_per_task_all_models}--\ref{tab:full_per_task_ret} provide the complete score matrix for all 36 models across every \mteblm{} task, presented one table per category so that each fits the page upright.
The tables share a common layout: models are rows, ordered by overall \mteblm{} score, with the ten LLMs listed first and the 26 embedding models below; the category's tasks are columns, with its metric given in the caption.
The final column gives each model's mean over the category, matching the corresponding column of Table~\ref{tab:main_results}.
Bold marks the best value in a column, whether a single task or the category mean.
Because rows are ordered by overall score rather than by category, this column is not monotonic: it shows where a model over- or under-performs its overall rank.
These tables extend Table~\ref{tab:per_task_scores} to all models.

\begin{table*}[htbp]
\centering
\small
\setlength{\tabcolsep}{4pt}
\renewcommand{\arraystretch}{0.95}
\resizebox{\ifdim\width>\textwidth\textwidth\else\width\fi}{!}{%
\begin{tabular}{@{}lrrrrrrrr@{\hspace{6pt}}!{\color{gray!35}\vrule width 0.3pt}@{\hspace{6pt}}r@{}}
\toprule
\textbf{Model} & \textbf{AmazonCF} & \textbf{Banking77} & \textbf{IMDB} & \textbf{MTOPDomain} & \textbf{MassiveIntent} & \textbf{MassiveScenario} & \textbf{ToxicConvs} & \textbf{TweetSent} & \textbf{Mean} \\
\midrule
\multicolumn{10}{l}{\emph{LLMs}} \\
Gemini 3.1 Pro & 84.8 & 85.0 & \textbf{98.0} & 95.5 & 84.9 & 79.4 & 89.6 & 64.2 & 85.2 \\
Gemini 3 Flash & 81.2 & 83.1 & 97.6 & 96.4 & 85.4 & 75.9 & 90.0 & 63.0 & 84.1 \\
Qwen3.6-27B & 90.1 & 80.5 & 97.4 & 96.9 & 84.6 & 76.4 & 89.2 & 63.2 & 84.8 \\
Qwen3.6-35B-A3B & 88.7 & 77.8 & 97.4 & 96.8 & 79.8 & 76.6 & 91.0 & 62.4 & 83.8 \\
Kimi-K2.6 & 89.3 & 82.8 & 97.4 & 96.6 & 83.2 & 76.5 & 88.8 & 61.6 & 84.5 \\
DeepSeek-V4-Flash & 82.4 & 78.5 & 96.8 & 95.8 & 82.4 & 77.5 & 80.4 & 61.2 & 81.9 \\
MiniMax-M2.7 & 83.6 & 78.4 & 97.0 & 95.2 & 75.1 & 67.0 & 87.6 & 63.0 & 80.9 \\
GLM-4.7 & 88.7 & 80.4 & 97.2 & 96.8 & 83.0 & 77.1 & 87.6 & 63.2 & 84.2 \\
Gemini 3.1 FLite & 79.7 & 78.7 & 97.6 & 95.9 & 83.2 & 76.2 & 88.6 & 63.4 & 82.9 \\
DeepSeek-R1 & 86.9 & 78.9 & 97.0 & 96.1 & 82.7 & 76.6 & 78.2 & 63.4 & 82.5 \\
\midrule
\multicolumn{10}{l}{\emph{Embedding models}} \\
Octen-8B & \textbf{93.0} & 87.3 & 97.7 & 98.2 & 85.7 & 89.3 & 91.9 & 77.9 & 90.1 \\
Qwen3-E-8B & 92.9 & 87.2 & 97.8 & 98.1 & 85.8 & 89.4 & 92.0 & 77.8 & 90.1 \\
Qwen3-E-4B & 92.6 & 86.3 & 97.4 & 97.6 & 85.2 & 86.0 & 92.3 & 77.1 & 89.3 \\
Nemotron-8B & 83.8 & 83.6 & 97.1 & 96.5 & 83.5 & 83.5 & 88.3 & 62.1 & 84.8 \\
KaLM-12B & 90.4 & 87.7 & 96.3 & 98.5 & 84.6 & 86.7 & 90.2 & 76.1 & 88.8 \\
Jina-v5-S & 91.8 & 91.5 & 95.9 & \textbf{99.2} & \textbf{88.9} & 92.6 & \textbf{94.4} & 68.5 & 90.4 \\
SFR-2 & 92.7 & 90.1 & 97.6 & 98.3 & 86.3 & 90.5 & 91.9 & \textbf{78.9} & \textbf{90.8} \\
Jina-v5-Nano & 91.3 & 90.2 & 95.5 & 98.0 & 88.4 & \textbf{93.0} & 94.4 & 66.3 & 89.6 \\
F2LLM-14B & 64.1 & 84.7 & 91.2 & 98.8 & 77.6 & 89.4 & 58.7 & 56.7 & 77.7 \\
GTE-Qwen2-7B & 86.8 & 84.9 & 96.7 & 98.0 & 83.7 & 85.7 & 88.2 & 70.0 & 86.7 \\
Linq-Mistral & 83.9 & 87.7 & 94.9 & 97.0 & 82.7 & 84.5 & 70.2 & 59.2 & 82.5 \\
Qwen3-E-0.6B & 90.5 & 80.8 & 96.3 & 95.8 & 80.1 & 83.8 & 81.9 & 74.3 & 85.4 \\
EmbGemma-300M & 89.6 & \textbf{91.6} & 91.9 & 98.6 & 85.6 & 91.6 & 83.4 & 59.8 & 86.5 \\
F2LLM-8B & 63.1 & 83.6 & 87.3 & 99.0 & 72.2 & 85.8 & 59.3 & 51.4 & 75.2 \\
F2LLM-4B & 61.3 & 82.2 & 86.4 & 99.0 & 72.3 & 86.5 & 60.7 & 54.2 & 75.3 \\
F2LLM-1.7B & 61.2 & 77.6 & 83.4 & 98.3 & 74.6 & 87.5 & 61.1 & 54.3 & 74.7 \\
GTE-Qwen2-1.5B & 81.6 & 79.7 & 95.8 & 95.3 & 77.9 & 79.8 & 84.9 & 71.3 & 83.3 \\
F2LLM-0.6B & 59.5 & 73.8 & 79.7 & 97.3 & 73.4 & 87.3 & 59.7 & 49.0 & 72.5 \\
mE5-L-Inst & 69.1 & 77.4 & 96.4 & 91.6 & 70.0 & 69.1 & 64.8 & 56.8 & 74.4 \\
BGE-M3 & 75.4 & 82.4 & 87.9 & 93.5 & 72.2 & 76.6 & 64.8 & 56.9 & 76.2 \\
mE5-L & 79.1 & 75.3 & 89.2 & 91.4 & 68.7 & 72.8 & 58.1 & 51.8 & 73.3 \\
Arctic-L-v2 & 63.6 & 82.3 & 73.4 & 93.5 & 72.2 & 76.0 & 61.5 & 48.8 & 71.4 \\
mE5-B & 79.6 & 73.7 & 85.5 & 90.6 & 66.1 & 71.3 & 57.1 & 51.6 & 71.9 \\
mE5-S & 71.8 & 70.6 & 79.7 & 89.1 & 65.6 & 69.1 & 59.4 & 52.9 & 69.8 \\
E5-Mistral-7B & 71.8 & 71.4 & 91.6 & 79.3 & 65.8 & 73.4 & 62.2 & 44.9 & 70.0 \\
GritLM-7B & 74.7 & 70.6 & 79.0 & 84.9 & 65.2 & 68.3 & 61.4 & 45.0 & 68.6 \\
\bottomrule
\end{tabular}
}
\caption{\textbf{Per-task scores: Classification (Accuracy).}}
\label{tab:full_per_task_all_models}
\end{table*}

\begin{table*}[htbp]
\centering
\small
\setlength{\tabcolsep}{4pt}
\renewcommand{\arraystretch}{0.95}
\resizebox{\ifdim\width>\textwidth\textwidth\else\width\fi}{!}{%
\begin{tabular}{@{}lrrrrrrrrrr@{\hspace{6pt}}!{\color{gray!35}\vrule width 0.3pt}@{\hspace{6pt}}r@{}}
\toprule
\textbf{Model} & \textbf{BIOSSES} & \textbf{SICK-R} & \textbf{STS12} & \textbf{STS13} & \textbf{STS14} & \textbf{STS15} & \textbf{STS16} & \textbf{STS17} & \textbf{STS22v2} & \textbf{STSBench} & \textbf{Mean} \\
\midrule
\multicolumn{12}{l}{\emph{LLMs}} \\
Gemini 3.1 Pro & 89.6 & 86.1 & 82.6 & 91.9 & 90.5 & 94.4 & 89.0 & \textbf{94.2} & \textbf{73.6} & 92.6 & 88.5 \\
Gemini 3 Flash & 88.7 & 86.8 & 79.5 & 91.5 & 89.3 & 93.9 & 88.5 & 94.1 & 71.7 & 91.7 & 87.6 \\
Qwen3.6-27B & 85.0 & 76.4 & 77.1 & 90.6 & 88.0 & 91.8 & 87.1 & 91.8 & 72.0 & 89.6 & 84.9 \\
Qwen3.6-35B-A3B & 84.0 & 75.5 & 75.1 & 89.6 & 85.1 & 91.4 & 85.1 & 92.7 & 73.4 & 87.8 & 84.0 \\
Kimi-K2.6 & 86.4 & 82.1 & 72.2 & 89.3 & 85.1 & 91.1 & 83.9 & 90.1 & 70.9 & 87.8 & 83.9 \\
DeepSeek-V4-Flash & 78.5 & 75.3 & 75.6 & 87.3 & 84.5 & 89.2 & 82.4 & 91.5 & 69.2 & 87.9 & 82.2 \\
MiniMax-M2.7 & 87.7 & 79.7 & 65.4 & 86.5 & 81.9 & 88.4 & 81.1 & 87.1 & 66.7 & 85.7 & 81.0 \\
GLM-4.7 & 82.3 & 76.0 & 73.1 & 88.5 & 85.7 & 91.9 & 84.9 & 90.4 & 73.2 & 88.5 & 83.5 \\
Gemini 3.1 FLite & \textbf{89.8} & 78.0 & 76.0 & 90.5 & 86.3 & 91.4 & 86.5 & 91.4 & 71.2 & 90.3 & 85.1 \\
DeepSeek-R1 & 85.0 & 79.7 & 71.9 & 88.7 & 85.6 & 89.9 & 84.2 & 91.5 & 71.4 & 89.2 & 83.7 \\
\midrule
\multicolumn{12}{l}{\emph{Embedding models}} \\
Octen-8B & 84.4 & 87.9 & 87.3 & 93.9 & 90.9 & 94.3 & 92.5 & 93.6 & 69.5 & 93.2 & 88.7 \\
Qwen3-E-8B & 82.0 & 88.4 & 87.3 & 94.0 & 91.0 & 94.2 & 92.5 & 93.7 & 68.9 & 93.4 & 88.5 \\
Qwen3-E-4B & 84.8 & 87.7 & \textbf{87.4} & \textbf{94.6} & \textbf{91.4} & \textbf{94.5} & \textbf{92.9} & 93.2 & 67.7 & 93.7 & \textbf{88.8} \\
Nemotron-8B & 86.3 & 85.4 & 83.4 & 91.5 & 87.9 & 92.1 & 88.7 & 91.4 & 64.3 & 90.8 & 86.2 \\
KaLM-12B & 87.0 & 81.9 & 81.4 & 89.2 & 85.9 & 90.6 & 86.7 & 89.0 & 65.1 & 89.1 & 84.6 \\
Jina-v5-S & 85.3 & 90.4 & 86.6 & 89.2 & 88.9 & 92.6 & 87.4 & 89.1 & 65.0 & \textbf{94.9} & 86.9 \\
SFR-2 & 87.5 & 76.7 & 74.6 & 81.4 & 80.6 & 87.5 & 84.6 & 85.0 & 38.2 & 82.6 & 77.9 \\
Jina-v5-Nano & 87.4 & \textbf{91.1} & 86.2 & 89.9 & 89.2 & 93.1 & 85.4 & 88.4 & 64.9 & 94.2 & 87.0 \\
F2LLM-14B & 88.2 & 82.8 & 82.4 & 86.9 & 84.8 & 91.3 & 85.6 & 88.9 & 63.4 & 87.0 & 84.1 \\
GTE-Qwen2-7B & 83.0 & 79.7 & 78.1 & 88.6 & 83.4 & 89.5 & 85.5 & 87.1 & 54.7 & 86.2 & 81.6 \\
Linq-Mistral & 86.1 & 84.4 & 77.6 & 87.8 & 84.1 & 91.2 & 87.4 & 89.9 & 56.1 & 89.0 & 83.4 \\
Qwen3-E-0.6B & 84.6 & 84.6 & 83.6 & 92.1 & 87.0 & 91.7 & 89.7 & 87.9 & 64.6 & 91.3 & 85.7 \\
EmbGemma-300M & 83.2 & 81.9 & 79.1 & 85.5 & 84.5 & 89.2 & 84.7 & 86.6 & 57.3 & 88.8 & 82.1 \\
F2LLM-8B & 88.8 & 82.4 & 81.3 & 86.6 & 83.8 & 91.3 & 85.4 & 89.3 & 63.2 & 86.5 & 83.9 \\
F2LLM-4B & 87.5 & 81.9 & 82.3 & 86.1 & 83.8 & 91.5 & 85.2 & 87.9 & 62.8 & 87.1 & 83.6 \\
F2LLM-1.7B & 88.5 & 81.3 & 82.6 & 86.7 & 84.5 & 91.3 & 85.2 & 87.8 & 62.9 & 87.6 & 83.8 \\
GTE-Qwen2-1.5B & 81.4 & 80.7 & 72.9 & 84.4 & 81.7 & 88.7 & 84.6 & 86.1 & 56.6 & 85.6 & 80.3 \\
F2LLM-0.6B & 87.8 & 80.0 & 82.7 & 86.7 & 84.1 & 90.6 & 84.9 & 84.0 & 62.3 & 87.6 & 83.1 \\
mE5-L-Inst & 87.0 & 81.8 & 83.5 & 86.8 & 85.1 & 91.8 & 86.6 & 87.0 & 60.8 & 89.0 & 83.9 \\
BGE-M3 & 83.4 & 79.3 & 79.8 & 78.6 & 80.7 & 88.3 & 84.7 & 82.3 & 60.5 & 86.0 & 80.4 \\
mE5-L & 84.6 & 80.5 & 80.9 & 75.6 & 78.7 & 89.7 & 83.6 & 85.4 & 58.5 & 85.8 & 80.3 \\
Arctic-L-v2 & 87.2 & 72.5 & 71.6 & 79.6 & 76.0 & 83.9 & 83.2 & 79.4 & 59.6 & 78.4 & 77.2 \\
mE5-B & 86.2 & 78.9 & 79.1 & 75.9 & 79.2 & 88.7 & 82.0 & 81.6 & 56.0 & 84.5 & 79.2 \\
mE5-S & 84.4 & 78.0 & 78.3 & 77.9 & 77.9 & 87.9 & 82.6 & 77.0 & 56.8 & 84.2 & 78.5 \\
E5-Mistral-7B & 54.7 & 55.7 & 45.9 & 52.1 & 43.4 & 50.2 & 79.0 & 62.5 & -21.4 & 75.5 & 49.8 \\
GritLM-7B & 65.3 & 60.5 & 51.3 & 61.1 & 56.3 & 69.3 & 66.1 & 36.0 & 31.6 & 52.9 & 55.1 \\
\bottomrule
\end{tabular}
}
\caption{\textbf{Per-task scores: STS (Spearman $\rho$).}}
\label{tab:full_per_task_sts}
\end{table*}

\begin{table*}[htbp]
\centering
\small
\setlength{\tabcolsep}{4pt}
\renewcommand{\arraystretch}{0.95}
\resizebox{\ifdim\width>\textwidth\textwidth\else\width\fi}{!}{%
\begin{tabular}{@{}lrrrrrrrrr@{\hspace{6pt}}!{\color{gray!35}\vrule width 0.3pt}@{\hspace{6pt}}r@{}}
\toprule
\textbf{Model} & \textbf{ArxivP2P} & \textbf{ArxivS2S} & \textbf{BioP2P} & \textbf{MedP2P} & \textbf{MedS2S} & \textbf{Reddit} & \textbf{SE-P2P} & \textbf{SE-Cl} & \textbf{20News} & \textbf{Mean} \\
\midrule
\multicolumn{11}{l}{\emph{LLMs}} \\
Gemini 3.1 Pro & 60.1 & \textbf{62.5} & 62.5 & 52.5 & 54.0 & \textbf{93.9} & 42.9 & \textbf{91.8} & 78.7 & 66.6 \\
Gemini 3 Flash & 60.3 & 60.1 & 60.8 & 50.6 & 50.3 & 90.5 & 49.0 & 90.2 & \textbf{79.9} & 65.7 \\
Qwen3.6-27B & 54.5 & 41.6 & 49.9 & 48.0 & 42.7 & 60.1 & 46.1 & 73.6 & 66.3 & 53.6 \\
Qwen3.6-35B-A3B & 48.5 & 60.4 & 53.2 & 49.4 & 46.2 & 69.2 & 42.2 & 72.7 & 55.7 & 55.3 \\
Kimi-K2.6 & 52.6 & 44.2 & 51.9 & 50.6 & 39.1 & 57.4 & 52.1 & 81.8 & 68.9 & 55.4 \\
DeepSeek-V4-Flash & 37.4 & 45.5 & 29.3 & 27.0 & 40.9 & 55.2 & 40.3 & 53.5 & 64.4 & 43.7 \\
MiniMax-M2.7 & 36.4 & 40.6 & 38.3 & 39.8 & 42.4 & 67.4 & 39.4 & 75.1 & 57.9 & 48.6 \\
GLM-4.7 & 23.6 & 28.6 & 33.2 & 37.5 & 37.8 & 40.6 & 37.9 & 49.5 & 39.8 & 36.5 \\
Gemini 3.1 FLite & 14.7 & 18.9 & 17.4 & 19.7 & 21.6 & 28.5 & 21.5 & 25.8 & 27.2 & 21.7 \\
DeepSeek-R1 & 22.0 & 25.3 & 32.4 & 31.7 & 33.4 & 26.6 & 33.1 & 42.1 & 34.9 & 31.3 \\
\midrule
\multicolumn{11}{l}{\emph{Embedding models}} \\
Octen-8B & 61.9 & 60.5 & 61.6 & 55.9 & 53.3 & 78.7 & 59.7 & 85.8 & 68.5 & 65.1 \\
Qwen3-E-8B & 62.5 & 61.1 & 65.3 & 55.9 & 55.2 & 79.8 & 59.5 & 86.3 & 67.6 & 65.9 \\
Qwen3-E-4B & 62.6 & 60.3 & 64.0 & 55.5 & 54.1 & 77.1 & 58.8 & 83.9 & 65.3 & 64.6 \\
Nemotron-8B & 59.2 & 59.1 & 65.2 & 52.9 & 52.6 & 79.7 & 58.7 & 84.6 & 65.2 & 64.1 \\
KaLM-12B & 61.6 & 57.4 & 64.1 & 54.7 & 53.2 & 76.1 & 58.5 & 83.3 & 62.0 & 63.4 \\
Jina-v5-S & 60.0 & 57.5 & 63.6 & 53.7 & 52.6 & 70.5 & 58.9 & 76.8 & 59.9 & 61.5 \\
SFR-2 & 61.8 & 59.6 & 65.6 & 58.9 & \textbf{57.5} & 80.8 & 59.3 & 83.2 & 73.8 & \textbf{66.7} \\
Jina-v5-Nano & 59.1 & 55.4 & 61.6 & 54.6 & 52.1 & 67.4 & 57.3 & 76.8 & 59.7 & 60.4 \\
F2LLM-14B & 63.1 & 60.6 & 75.6 & 57.7 & 56.7 & 77.9 & 57.2 & 81.1 & 68.8 & 66.5 \\
GTE-Qwen2-7B & \textbf{64.1} & 61.5 & 66.7 & 57.0 & 54.1 & 80.5 & \textbf{60.7} & 84.5 & 58.1 & 65.2 \\
Linq-Mistral & 58.0 & 55.7 & 61.3 & 51.5 & 52.8 & 77.0 & 56.2 & 78.4 & 61.1 & 61.3 \\
Qwen3-E-0.6B & 60.2 & 57.6 & 61.8 & 52.9 & 51.1 & 69.9 & 56.0 & 77.3 & 61.7 & 60.9 \\
EmbGemma-300M & 57.9 & 53.9 & 62.8 & 54.4 & 51.0 & 75.9 & 51.9 & 73.7 & 51.8 & 59.3 \\
F2LLM-8B & 61.7 & 60.9 & 75.5 & 59.2 & 54.4 & 75.3 & 58.2 & 81.2 & 65.8 & 65.8 \\
F2LLM-4B & 61.3 & 59.2 & 74.2 & 59.8 & 54.7 & 74.4 & 55.2 & 79.9 & 63.9 & 64.7 \\
F2LLM-1.7B & 62.6 & 59.6 & \textbf{77.3} & \textbf{60.7} & 56.1 & 74.4 & 55.2 & 75.5 & 64.2 & 65.1 \\
GTE-Qwen2-1.5B & 58.5 & 57.5 & 62.1 & 53.2 & 52.3 & 71.1 & 52.3 & 73.6 & 59.1 & 60.0 \\
F2LLM-0.6B & 61.7 & 58.6 & 71.1 & 58.7 & 54.5 & 69.7 & 56.0 & 76.0 & 59.9 & 62.9 \\
mE5-L-Inst & 56.0 & 53.0 & 57.8 & 51.9 & 51.2 & 71.7 & 53.3 & 76.9 & 60.2 & 59.1 \\
BGE-M3 & 41.7 & 32.4 & 49.7 & 43.1 & 40.7 & 61.2 & 42.6 & 56.3 & 45.5 & 45.9 \\
mE5-L & 45.8 & 39.7 & 52.7 & 43.2 & 43.9 & 65.6 & 43.7 & 56.4 & 43.9 & 48.3 \\
Arctic-L-v2 & 48.1 & 38.4 & 48.7 & 46.5 & 44.0 & 65.2 & 46.6 & 61.7 & 48.3 & 49.7 \\
mE5-B & 44.9 & 40.2 & 46.2 & 43.1 & 44.3 & 64.8 & 42.6 & 57.6 & 43.4 & 47.5 \\
mE5-S & 42.5 & 39.9 & 48.2 & 45.5 & 44.3 & 65.5 & 41.9 & 58.0 & 43.2 & 47.7 \\
E5-Mistral-7B & 51.2 & 44.6 & 49.9 & 45.6 & 40.0 & 54.7 & 37.3 & 65.3 & 42.4 & 47.9 \\
GritLM-7B & 55.7 & 40.4 & 51.7 & 46.5 & 42.7 & 63.9 & 43.9 & 59.8 & 35.5 & 48.9 \\
\bottomrule
\end{tabular}
}
\caption{\textbf{Per-task scores: Clustering (V-measure).}}
\label{tab:full_per_task_clu}
\end{table*}

\begin{table*}[htbp]
\centering
\small
\setlength{\tabcolsep}{4pt}
\renewcommand{\arraystretch}{0.95}
\resizebox{\ifdim\width>\textwidth\textwidth\else\width\fi}{!}{%
\begin{tabular}{@{}lrrrr@{\hspace{6pt}}!{\color{gray!35}\vrule width 0.3pt}@{\hspace{6pt}}r@{}}
\toprule
\textbf{Model} & \textbf{LegalPC} & \textbf{RTE3} & \textbf{SprintDup} & \textbf{TwtURL} & \textbf{Mean} \\
\midrule
\multicolumn{6}{l}{\emph{LLMs}} \\
Gemini 3.1 Pro & 72.2 & 95.4 & 82.2 & 82.8 & 83.2 \\
Gemini 3 Flash & \textbf{77.4} & \textbf{96.7} & 83.4 & 87.6 & 86.3 \\
Qwen3.6-27B & 71.6 & 92.9 & 84.6 & 85.2 & 83.6 \\
Qwen3.6-35B-A3B & 69.4 & 93.6 & 85.2 & 83.4 & 82.9 \\
Kimi-K2.6 & 69.2 & 88.8 & 78.8 & 75.4 & 78.0 \\
DeepSeek-V4-Flash & 67.2 & 90.4 & 86.6 & 82.0 & 81.6 \\
MiniMax-M2.7 & 71.8 & 91.5 & 82.8 & 85.0 & 82.8 \\
GLM-4.7 & 69.6 & 90.0 & 82.2 & 76.8 & 79.7 \\
Gemini 3.1 FLite & 72.4 & 95.0 & 86.6 & 80.8 & 83.7 \\
DeepSeek-R1 & 68.8 & 90.0 & 79.6 & 79.6 & 79.5 \\
\midrule
\multicolumn{6}{l}{\emph{Embedding models}} \\
Octen-8B & 70.0 & 85.4 & 99.4 & 89.4 & 86.1 \\
Qwen3-E-8B & 70.4 & 85.7 & 99.6 & \textbf{90.2} & 86.5 \\
Qwen3-E-4B & 72.0 & 84.8 & \textbf{99.8} & 89.4 & 86.5 \\
Nemotron-8B & 72.6 & 84.8 & 99.4 & 89.8 & 86.7 \\
KaLM-12B & 73.6 & 85.2 & \textbf{99.8} & 89.6 & \textbf{87.1} \\
Jina-v5-S & 68.0 & 85.2 & 99.6 & 87.4 & 85.1 \\
SFR-2 & 67.4 & 84.8 & \textbf{99.8} & 89.0 & 85.3 \\
Jina-v5-Nano & 68.8 & 85.0 & 99.4 & 86.6 & 85.0 \\
F2LLM-14B & 72.0 & 84.8 & 99.6 & 88.6 & 86.3 \\
GTE-Qwen2-7B & 71.2 & 85.2 & 99.6 & 88.0 & 86.0 \\
Linq-Mistral & 71.0 & 84.8 & \textbf{99.8} & 88.6 & 86.1 \\
Qwen3-E-0.6B & 70.4 & 84.8 & \textbf{99.8} & 89.4 & 86.1 \\
EmbGemma-300M & 71.0 & 84.8 & 99.6 & 88.6 & 86.0 \\
F2LLM-8B & 72.0 & 84.8 & 99.6 & 88.0 & 86.1 \\
F2LLM-4B & 70.6 & 84.8 & \textbf{99.8} & 88.4 & 85.9 \\
F2LLM-1.7B & 71.0 & 84.8 & 99.6 & 89.0 & 86.1 \\
GTE-Qwen2-1.5B & 73.0 & 84.8 & 99.4 & 89.2 & 86.6 \\
F2LLM-0.6B & 70.8 & 85.0 & 99.4 & 88.0 & 85.8 \\
mE5-L-Inst & 70.8 & 84.8 & 99.6 & 89.2 & 86.1 \\
BGE-M3 & 71.6 & 84.8 & 99.6 & 87.6 & 85.9 \\
mE5-L & 64.0 & 85.0 & 98.4 & 88.2 & 83.9 \\
Arctic-L-v2 & 64.2 & 84.8 & 99.6 & 86.8 & 83.9 \\
mE5-B & 65.6 & 84.8 & 98.2 & 88.6 & 84.3 \\
mE5-S & 62.2 & 85.2 & 99.2 & 89.0 & 83.9 \\
E5-Mistral-7B & 66.0 & 84.8 & 73.4 & 67.4 & 72.9 \\
GritLM-7B & 56.8 & 84.8 & 92.0 & 75.4 & 77.3 \\
\bottomrule
\end{tabular}
}
\caption{\textbf{Per-task scores: Pair Classification (AP / Accuracy).}}
\label{tab:full_per_task_pair}
\end{table*}

\begin{table*}[htbp]
\centering
\small
\setlength{\tabcolsep}{4pt}
\renewcommand{\arraystretch}{0.95}
\resizebox{\ifdim\width>\textwidth\textwidth\else\width\fi}{!}{%
\begin{tabular}{@{}lrrrrrr@{\hspace{6pt}}!{\color{gray!35}\vrule width 0.3pt}@{\hspace{6pt}}r@{}}
\toprule
\textbf{Model} & \textbf{AILA} & \textbf{FQuAD} & \textbf{HC3Fin} & \textbf{ConsumerQA} & \textbf{PubHealth} & \textbf{TwtHjerne} & \textbf{Mean} \\
\midrule
\multicolumn{8}{l}{\emph{LLMs}} \\
Gemini 3.1 Pro & 14.5 & 92.0 & \textbf{71.0} & 86.0 & \textbf{90.0} & \textbf{33.4} & \textbf{64.5} \\
Gemini 3 Flash & 5.7 & 88.0 & 60.0 & 79.0 & 49.0 & 32.4 & 52.3 \\
Qwen3.6-27B & 14.2 & \textbf{96.0} & 57.0 & \textbf{88.0} & 86.0 & 33.0 & 62.4 \\
Qwen3.6-35B-A3B & 12.9 & 95.0 & 55.0 & 86.0 & 83.0 & 30.5 & 60.4 \\
Kimi-K2.6 & 13.2 & 91.0 & 56.0 & 80.0 & 67.0 & 33.0 & 56.7 \\
DeepSeek-V4-Flash & 13.3 & 83.0 & 50.0 & 76.0 & 64.0 & \textbf{33.4} & 53.3 \\
MiniMax-M2.7 & 9.7 & 82.0 & 42.0 & 67.0 & 59.0 & 28.2 & 48.0 \\
GLM-4.7 & 13.0 & 78.0 & 56.0 & 78.0 & 70.0 & 31.6 & 54.4 \\
Gemini 3.1 FLite & 10.2 & 84.0 & 46.0 & 77.0 & 45.0 & 31.9 & 49.0 \\
DeepSeek-R1 & 11.9 & 57.0 & 43.0 & 62.0 & 64.0 & 30.6 & 44.7 \\
\midrule
\multicolumn{8}{l}{\emph{Embedding models}} \\
Octen-8B & \textbf{23.2} & 69.0 & 66.0 & 67.0 & 81.0 & 29.9 & 56.0 \\
Qwen3-E-8B & 21.8 & 68.0 & 59.0 & 70.0 & 78.0 & 28.6 & 54.2 \\
Qwen3-E-4B & 21.0 & 61.0 & 56.0 & 64.0 & 75.0 & 24.8 & 50.3 \\
Nemotron-8B & 9.1 & 72.0 & 67.0 & 70.0 & 80.0 & 29.9 & 54.7 \\
KaLM-12B & 10.3 & 70.0 & 55.0 & 63.0 & 71.0 & 30.2 & 49.9 \\
Jina-v5-S & 14.7 & 64.0 & 48.0 & 59.0 & 77.0 & 25.5 & 48.0 \\
SFR-2 & 11.2 & 67.0 & 57.0 & 56.0 & 74.0 & 27.6 & 48.8 \\
Jina-v5-Nano & 12.6 & 65.0 & 44.0 & 61.0 & 74.0 & 27.6 & 47.4 \\
F2LLM-14B & 12.2 & 63.0 & 67.0 & 54.0 & 85.0 & 31.5 & 52.1 \\
GTE-Qwen2-7B & 9.1 & 64.0 & 51.0 & 55.0 & 77.0 & 19.0 & 45.8 \\
Linq-Mistral & 10.6 & 68.0 & 54.0 & 66.0 & 78.0 & 29.1 & 51.0 \\
Qwen3-E-0.6B & 20.2 & 55.0 & 36.0 & 63.0 & 68.0 & 22.4 & 44.1 \\
EmbGemma-300M & 5.5 & 72.0 & 42.0 & 62.0 & 78.0 & 24.2 & 47.3 \\
F2LLM-8B & 8.9 & 63.0 & 63.0 & 51.0 & 85.0 & 28.9 & 50.0 \\
F2LLM-4B & 8.1 & 64.0 & 62.0 & 53.0 & 83.0 & 29.4 & 49.9 \\
F2LLM-1.7B & 5.4 & 66.0 & 55.0 & 54.0 & 82.0 & 28.8 & 48.5 \\
GTE-Qwen2-1.5B & 5.3 & 58.0 & 44.0 & 59.0 & 76.0 & 21.2 & 43.9 \\
F2LLM-0.6B & 3.6 & 56.0 & 52.0 & 54.0 & 80.0 & 26.7 & 45.4 \\
mE5-L-Inst & 6.5 & 68.0 & 44.0 & 50.0 & 73.0 & 27.6 & 44.8 \\
BGE-M3 & 7.4 & 62.0 & 40.0 & 58.0 & 74.0 & 24.2 & 44.3 \\
mE5-L & 4.9 & 69.0 & 40.0 & 51.0 & 73.0 & 23.9 & 43.6 \\
Arctic-L-v2 & 3.3 & 61.0 & 42.0 & 59.0 & 76.0 & 26.1 & 44.6 \\
mE5-B & 3.2 & 64.0 & 28.0 & 47.0 & 68.0 & 23.5 & 38.9 \\
mE5-S & 3.6 & 57.0 & 26.0 & 43.0 & 64.0 & 20.9 & 35.8 \\
E5-Mistral-7B & 2.7 & 29.0 & 21.0 & 17.0 & 47.0 & 7.1 & 20.6 \\
GritLM-7B & 1.0 & 1.0 & 3.0 & 8.0 & 18.0 & 5.3 & 6.1 \\
\bottomrule
\end{tabular}
}
\caption{\textbf{Per-task scores: Retrieval (Recall@1).}}
\label{tab:full_per_task_ret}
\end{table*}

\subsection{Benchmark Task Suite}
\label{app:datasets}

Table~\ref{tab:datasets} lists all 37 tasks with languages, sample counts, and source citations.
All are held-out subsets (seed\,=\,42) derived from MTEB and MMTEB tasks, hosted on Hugging Face under \texttt{mteb/llm-eval-*}; the exact dataset path and pinned revision for each task are listed in the released repository.
Table~\ref{tab:token_budget} provides per-task token budgets for cost estimation, broken down by split.

\begin{table}[ht]
\centering
\small
\setlength{\tabcolsep}{3pt}
\resizebox{\linewidth}{!}{%
\begin{tabular}{llrcll}
\toprule
\textbf{Task} & \textbf{Lang.} & \textbf{N} & \textbf{Cls.} & \textbf{Metric} & \textbf{Source} \\
\midrule
\addlinespace[4pt]
\multicolumn{6}{l}{\textit{Classification (8 tasks)}} \\
\addlinespace[2pt]
ImdbCls & en & 500 & 2 & Acc. & \citet{maas2011learning} \\
Banking77Cls & en & 3k & 77 & Acc. & \citet{casanueva2020banking77} \\
AmazonCounterfactualCls & en, de, ja & 809 & 2 & Acc. & \citet{oneill2021counterfactual} \\
MTOPDomainCls & en, de, fr & 2k & 11 & Acc. & \citet{li2021mtop} \\
MassiveIntentCls & en, de, fr, ja & 4k & 60 & Acc. & \citet{fitzgerald2022massive} \\
MassiveScenarioCls & en, de, fr, ja & 3k & 18 & Acc. & \citet{fitzgerald2022massive} \\
ToxicConversationsCls & en & 500 & 2 & Acc. & \citet{borkan2019nuanced} \\
TweetSentimentCls & en & 500 & 3 & Acc. & \citet{muennighoff2023mteb} \\
\addlinespace[4pt]
\multicolumn{6}{l}{\textit{Semantic Textual Similarity (10 tasks)}} \\
\addlinespace[2pt]
STSBenchmark & en & 500 & -- & Spearman & \citet{cer2017semeval} \\
SICK-R & en & 500 & -- & Spearman & \citet{marelli2014sick} \\
STS12 & en & 500 & -- & Spearman & \citet{agirre2012semeval} \\
STS13 & en & 500 & -- & Spearman & \citet{agirre2012semeval} \\
STS14 & en & 500 & -- & Spearman & \citet{agirre2012semeval} \\
STS15 & en & 500 & -- & Spearman & \citet{agirre2016semeval} \\
STS16 & en & 500 & -- & Spearman & \citet{agirre2016semeval} \\
BIOSSES & en & 100 & -- & Spearman & \citet{sougancioglu2017biosses} \\
STS17 & en, de, es, fr & 1k & -- & Spearman & \citet{cer2017semeval} \\
STS22v2 & en, de, es, fr, ru, zh & 2k & -- & Spearman & \citet{chen2022semeval} \\
\addlinespace[4pt]
\multicolumn{6}{l}{\textit{Clustering (9 tasks)}} \\
\addlinespace[2pt]
RedditClustP2P & en & 1k & -- & V-meas. & \citet{muennighoff2023mteb} \\
TwentyNewsgroupsV2 & en & 1k & -- & V-meas. & \citet{muennighoff2023mteb} \\
StackExchangeClustP2PV2 & en & 1k & -- & V-meas. & \citet{muennighoff2023mteb} \\
StackExchangeClustV2 & en & 1k & -- & V-meas. & \citet{muennighoff2023mteb} \\
ArxivClustP2P & en & 1k & -- & V-meas. & \citet{muennighoff2023mteb} \\
ArxivClustS2S & en & 1k & -- & V-meas. & \citet{muennighoff2023mteb} \\
BiorxivClustP2PV2 & en & 1k & -- & V-meas. & \citet{muennighoff2023mteb} \\
MedrxivClustP2PV2 & en & 1k & -- & V-meas. & \citet{muennighoff2023mteb} \\
MedrxivClustS2SV2 & en & 1k & -- & V-meas. & \citet{muennighoff2023mteb} \\
\addlinespace[4pt]
\multicolumn{6}{l}{\textit{Pair Classification (4 tasks)}} \\
\addlinespace[2pt]
SprintDuplicateQuestionsPC & en & 500 & 2 & Avg.~Prec. & \citet{shah2018adversarial} \\
TwitterURLCorpusPC & en & 500 & 2 & Avg.~Prec. & \citet{lan2017continuously} \\
LegalBenchPC & en & 500 & 2 & Avg.~Prec. & \citet{guha2023legalbench} \\
RTE3PC & de, en, fr, it & 2k & 2 & Avg.~Prec. & \citet{giampiccolo2007third} \\
\addlinespace[4pt]
\multicolumn{6}{l}{\textit{Retrieval (6 tasks)}} \\
\addlinespace[2pt]
AILAStatutes & en & 50\,Q / 82\,C & -- & Recall@1 & \citet{enevoldsen2025mmtebmassivemultilingualtext} \\
FQuADRetrieval & fr & 100\,Q / 269\,C & -- & Recall@1 & \citet{dhoffschmidt2020fquad} \\
HC3FinanceRetrieval & en & 100\,Q / 415\,C & -- & Recall@1 & \citet{guo2023hc3} \\
LegalBenchConsumerContractsQA & en & 100\,Q / 154\,C & -- & Recall@1 & \citet{guha2023legalbench} \\
PublicHealthQA & en & 100\,Q / 172\,C & -- & Recall@1 & HF dataset \\
TwitterHjerneRetrieval & da & 77\,Q / 262\,C & -- & Recall@1 & \citet{holm2024danoliterate} \\
\bottomrule
\end{tabular}
}
\caption{\textbf{\mteblm{} task suite (37 tasks).}
N = held-out test samples (summed over languages for multilingual tasks);
Q = queries, C = corpus documents.
Multilingual tasks are evaluated per language and averaged. Held-out subsets
(seed 42) derived from MTEB and MMTEB tasks \citep{muennighoff2023mteb,enevoldsen2025mmtebmassivemultilingualtext},
hosted at \texttt{mteb/llm-eval-*}. Token counts: Table~\ref{tab:token_budget}.}
\label{tab:datasets}
\end{table}

\begin{table}[ht]
\centering
\small
\setlength{\tabcolsep}{5pt}
\begin{tabular}{lrr}
\toprule
\multicolumn{3}{l}{\textit{Classification}} \\
\addlinespace[2pt]
\textbf{Task} & \textbf{Test} & \textbf{Train (kNN)\textsuperscript{*}} \\
\midrule
ImdbCls & 132k & 7,281k \\
Banking77Cls & 38k & 134k \\
AmazonCounterfactualCls & 22k & 258k \\
MTOPDomainCls & 23k & 393k \\
MassiveIntentCls & 37k & 433k \\
MassiveScenarioCls & 30k & 433k \\
ToxicConversationsCls & 33k & 3,239k \\
TweetSentimentCls & 9k & 483k \\
\addlinespace[4pt]
\multicolumn{3}{l}{\textit{STS}} \\
\addlinespace[2pt]
STSBenchmark & \multicolumn{2}{r}{12k} \\
SICK-R & \multicolumn{2}{r}{10k} \\
STS12 & \multicolumn{2}{r}{14k} \\
STS13 & \multicolumn{2}{r}{12k} \\
STS14 & \multicolumn{2}{r}{12k} \\
STS15 & \multicolumn{2}{r}{12k} \\
STS16 & \multicolumn{2}{r}{14k} \\
BIOSSES & \multicolumn{2}{r}{7k} \\
STS17 & \multicolumn{2}{r}{21k} \\
STS22v2 & \multicolumn{2}{r}{1,866k} \\
\addlinespace[4pt]
\multicolumn{3}{l}{\textit{Clustering}} \\
\addlinespace[2pt]
RedditClustP2P & \multicolumn{2}{r}{178k} \\
TwentyNewsgroupsV2 & \multicolumn{2}{r}{8k} \\
StackExchangeClustP2PV2 & \multicolumn{2}{r}{276k} \\
StackExchangeClustV2 & \multicolumn{2}{r}{13k} \\
ArxivClustP2P & \multicolumn{2}{r}{226k} \\
ArxivClustS2S & \multicolumn{2}{r}{16k} \\
BiorxivClustP2PV2 & \multicolumn{2}{r}{313k} \\
MedrxivClustP2PV2 & \multicolumn{2}{r}{401k} \\
MedrxivClustS2SV2 & \multicolumn{2}{r}{23k} \\
\addlinespace[4pt]
\multicolumn{3}{l}{\textit{PairClassification}} \\
\addlinespace[2pt]
SprintDuplicateQuestionsPC & \multicolumn{2}{r}{14k} \\
TwitterURLCorpusPC & \multicolumn{2}{r}{18k} \\
LegalBenchPC & \multicolumn{2}{r}{37k} \\
RTE3PC & \multicolumn{2}{r}{114k} \\
\midrule
\textbf{Task} & \textbf{Queries} & \textbf{Corpus} \\
\midrule
\multicolumn{3}{l}{\textit{Retrieval}} \\
\addlinespace[2pt]
AILAStatutes & 33k & 37k \\
FQuADRetrieval & 1k & 59k \\
HC3FinanceRetrieval & 1k & 89k \\
LegalBenchConsumerContractsQA & 2k & 83k \\
PublicHealthQA & 1k & 28k \\
TwitterHjerneRetrieval & 4k & 10k \\
\bottomrule
\end{tabular}
\caption{\textbf{Token budget per task} (GPT-4o tokenizer; raw text).
Actual counts vary by ${\pm}$15--40\% across model vocabularies.
\textsuperscript{*}Train split is the kNN reference corpus processed only by
embedding models; LLMs process only the test split. Corpus-in-context formatting
adds ${\sim}$20 tokens per document for LLM retrieval.}
\label{tab:token_budget}
\end{table}

\section{Ablation Studies}
\label{app:ablation_detail}

This section details the reduced-thinking and few-shot experiments and provides a per-task retrieval breakdown.

\subsection{Reduced-Thinking Ablation Details}
\label{app:ablation_reduced}

Our reduced-thinking ablation uses Gemini~3~Flash with \texttt{reasoning\_effort=low}, which instructs the model to minimise internal chain-of-thought reasoning.
We evaluate on all 6 retrieval tasks and 3 representative classification tasks (IMDB, Banking77, ToxicConversations).

The reduction differs by task type: 54--94\% on retrieval and 12--30\% on classification.
This suggests the model adaptively allocates more reasoning to retrieval (where documents must be read and compared) than classification (where the decision is more immediate).

Reduced reasoning preserves or improves all six Flash retrieval scores and four of six cross-family averages (Figure~\ref{fig:cost_breakdown}b).
We hypothesise that excessive reasoning introduces second-guessing: the model considers multiple interpretations of a document's relevance when a more direct reading would suffice.

On classification, all three tasks show $|\Delta| < 1.0$, indicating no measurable benefit from additional reasoning in this ablation.
This result is consistent with the supervision asymmetry between kNN and zero-shot classification.

\begin{table}[t]
\centering
\small
\resizebox{\linewidth}{!}{%
\begin{tabular}{lccccc}
\toprule
\textbf{Task} & \textbf{Gemini 3 Flash} & \textbf{Gemini 3 Flash (low)} & \textbf{$\Delta$} & \textbf{Think $\downarrow$} & \textbf{Best Emb.} \\
\midrule
\multicolumn{6}{l}{\emph{Retrieval}} \\
AILAStatutes & 5.7 & \textbf{12.0} & +6.3 & -- & 23.2 \\
FQuADRetrieval & 88.0 & \textbf{92.0} & +4.0 & 54\% & 72.0 \\
HC3FinanceRetrieval & 60.0 & \textbf{66.0} & +6.0 & 87\% & 67.0 \\
LegalBenchConsumerContractsQA & 79.0 & \textbf{83.0} & +4.0 & 85\% & 70.0 \\
PublicHealthQA & 49.0 & \textbf{66.0} & +17.0 & 94\% & 85.0 \\
TwitterHjerneRetrieval & 32.4 & \textbf{33.4} & +1.1 & -- & 31.5 \\
\bottomrule
\end{tabular}
}
\caption{\textbf{Reduced-thinking ablation} (Gemini 3 Flash with \texttt{reasoning\_effort=low} vs.\ default)
on the \mteblm{} retrieval tasks. Think~$\downarrow$ = reduction in thinking tokens vs.\ default.
Reducing thinking by 54--94\% improves all six retrieval scores in this ablation.}
\label{tab:ablation_nocot}
\end{table}

\begin{table}[t]
\centering
\small
\begin{tabular}{lcccccc}
\toprule
\textbf{Task} & \textbf{Classes} & \textbf{Zero-shot} & \textbf{5-shot} & \textbf{$\Delta$} & \textbf{Best Emb.} \\
\midrule
IMDB & 2 & \textbf{0.976} & 0.974 & $-$0.002 & \textbf{0.976} \\
ToxicConversations & 2 & \textbf{0.900} & 0.838 & $-$0.062 & 0.810 \\
TweetSentiment & 3 & 0.700 & 0.682 & $-$0.018 & \textbf{0.710} \\
Banking77 & 77 & 0.831 & 0.165 & $-$0.666 & \textbf{0.960} \\
\bottomrule
\end{tabular}
\caption{\textbf{Few-shot classification ablation} (Flash, 5 in-context examples vs.\ zero-shot).
\textbf{Bold} = best score per task across all methods.
Five-shot prompting matches zero-shot performance on the 2--3 class tasks and lowers Banking77 performance, where the prompt contains five examples for 77 labels.}
\label{tab:ablation_fewshot}
\end{table}

\subsection{Per-Task Retrieval Analysis}
\label{app:retrieval_detail}

The six retrieval tasks in our benchmark span qualitatively different retrieval challenges; per-task scores for all 36 models are in Table~\ref{tab:full_per_task_ret}.
We provide a per-task breakdown of the aggregate retrieval advantage ($+$8.5) reported in \S\ref{sec:categories}.
Pro leads five retrieval tasks, and Octen-8B leads AILAStatutes.

\paragraph{FQuAD (French reading-comprehension QA).}
Answering requires reading French passages in context and matching the supporting one.
Pro (92.0) leads the best embedding (72.0) by a wide margin, the largest LLM advantage in the suite.

\paragraph{Consumer-Contracts QA (legal).}
Yes/no questions over consumer-contract clauses.
Pro (86.0) outperforms the best embedding (70.0), consistent with the need to interpret the clause.

\paragraph{Public-Health QA.}
Pro (90.0) edges the best embedding (85.0); factual health QA is partly captured by dense retrieval, narrowing the gap.

\paragraph{HC3-Finance QA.}
Pro (71.0) narrowly leads the best embedding (67.0) on financial QA retrieval.

\paragraph{TwitterHjerne (Danish social media).}
Retrieving relevant tweets given a query.
Pro (33.4) slightly outperforms embeddings (best: 31.5); both handle keyword-centric retrieval similarly.

\paragraph{AILAStatutes (legal statute retrieval): where reasoning hurts.}
Matching case facts to relevant statutes.
The best embedding (Octen-8B, 23.2) outperforms Pro (14.5): Pro over-retrieves broadly associated provisions, while cosine similarity naturally produces a narrower, more precise ranking.

\section{Per-Category Rankings and Figures}
\label{app:category_rankings}

\begin{figure*}[htbp]
\centering
\includegraphics[width=\textwidth]{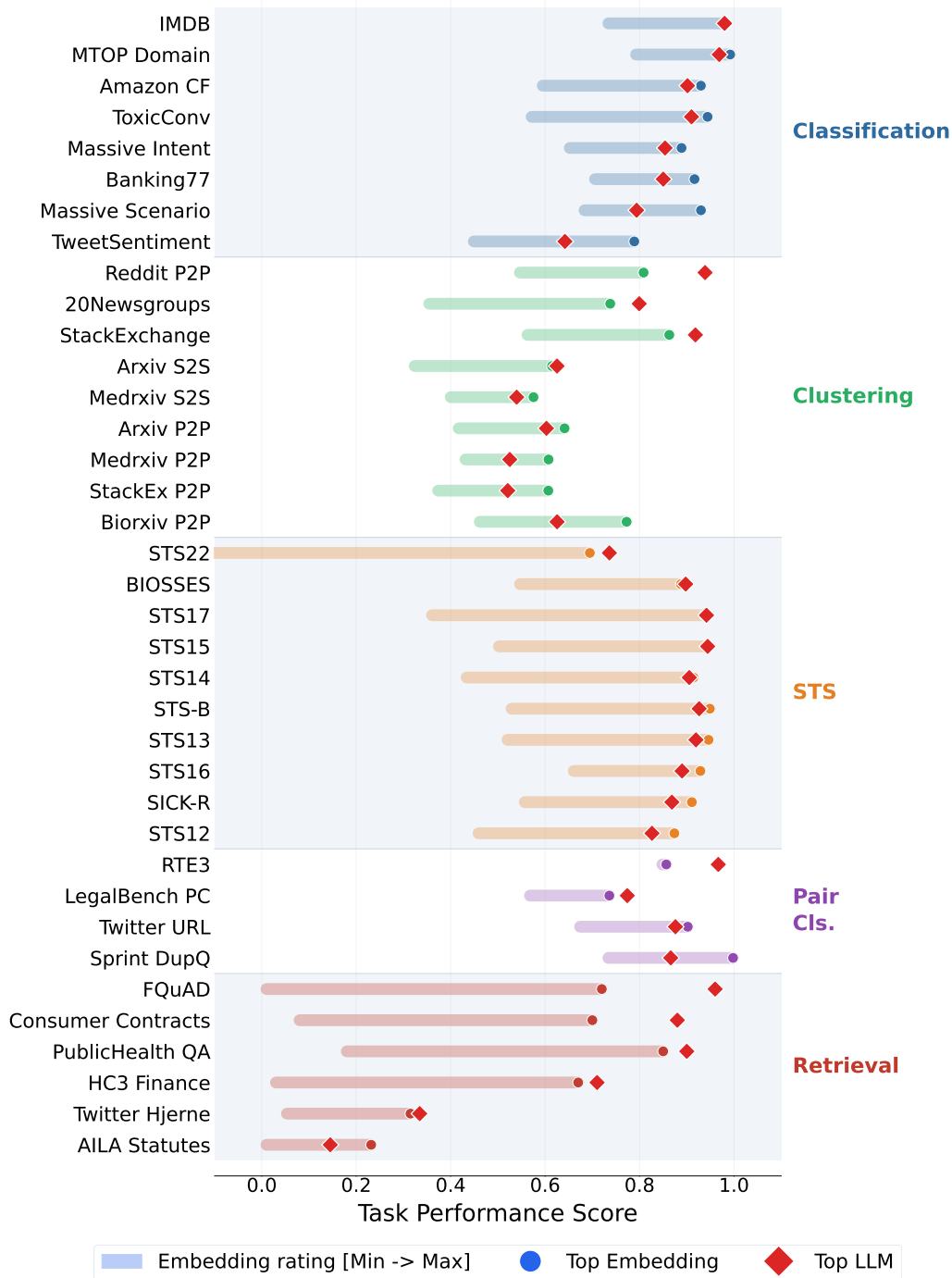}
\caption{\textbf{Per-task performance.}
Horizontal bars show the full embedding score range (min to max across 26 models) for each \mteblm{} task, grouped by category.
Circle = best embedding; diamond = best LLM on that task, taken over all ten (six different LLMs hold it across the suite, Gemini~3.1~Pro on 21 of 37 tasks).
Where the diamond falls inside the bar, some embedding model already matches the best LLM: this holds on 7 of 8 classification tasks and 7 of 10 STS tasks, but on only 1 of 6 retrieval tasks.}
\label{fig:taskrange}
\end{figure*}

This section provides task- and model-level figures, followed by category-level rankings.
Figure~\ref{fig:taskrange} is the task-level view of the paper's central claim and is referenced from \S\ref{sec:categories}.

\subsection{Task- and Model-Level Figures}
\label{app:figures}

The figures below show task-level performance, throughput, category profiles, and cross-model correlations.

\begin{figure}[htbp]
\centering
\includegraphics[width=\linewidth]{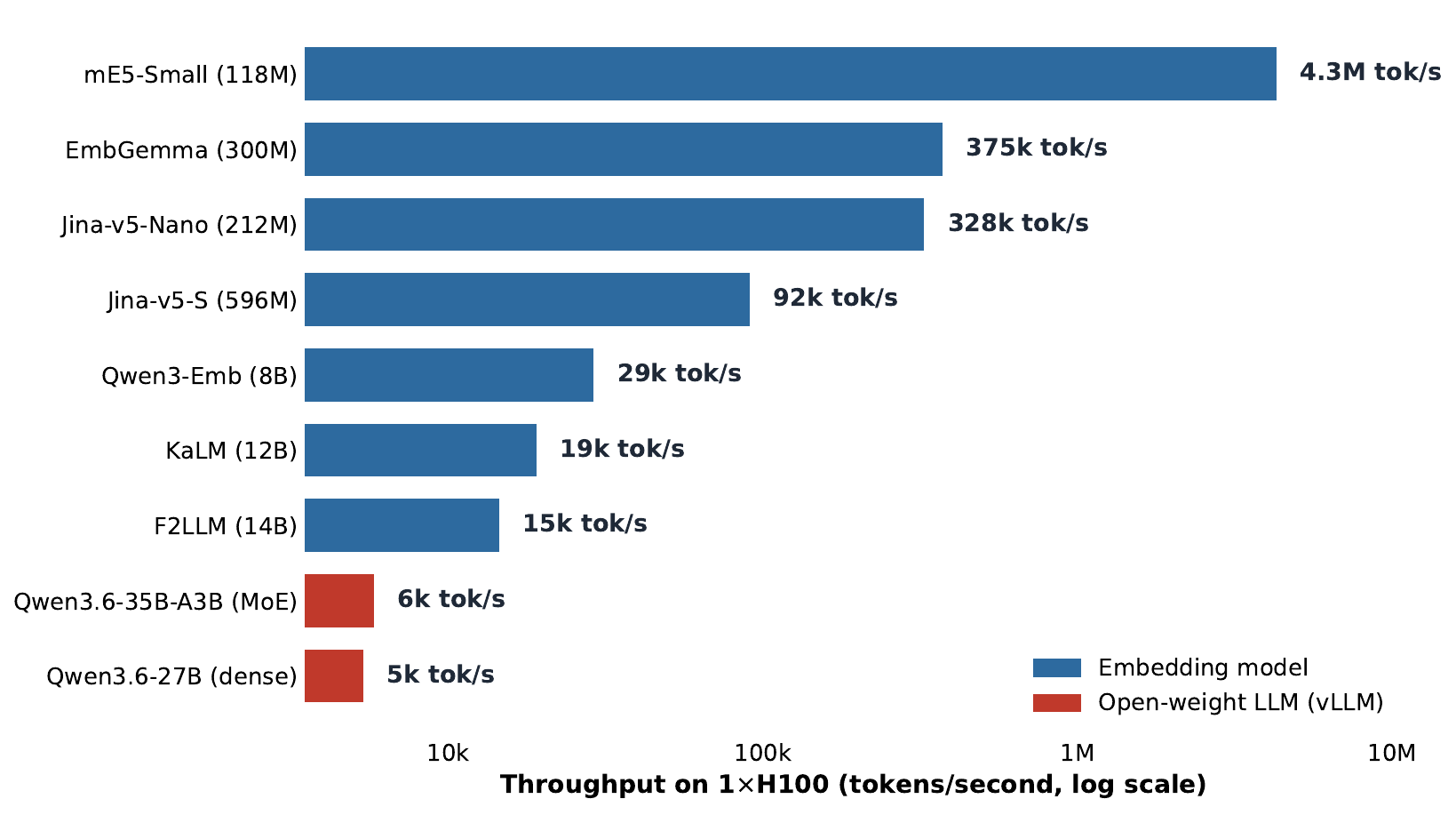}
\caption{\textbf{Same-hardware inference throughput.}
Two open-weight LLMs and seven representative embedding models (118M--14B) served on one H100 (tokens/second, log scale); the full 26-model embedding throughput is in Table~\ref{tab:embedding_throughput}.
Even the slowest embedding runs $\sim$2.5$\times$ faster than the fastest LLM; the fastest runs $\sim$736$\times$ faster.}
\label{fig:throughput}
\end{figure}

\begin{figure}[htbp]
\centering
\includegraphics[width=0.85\linewidth]{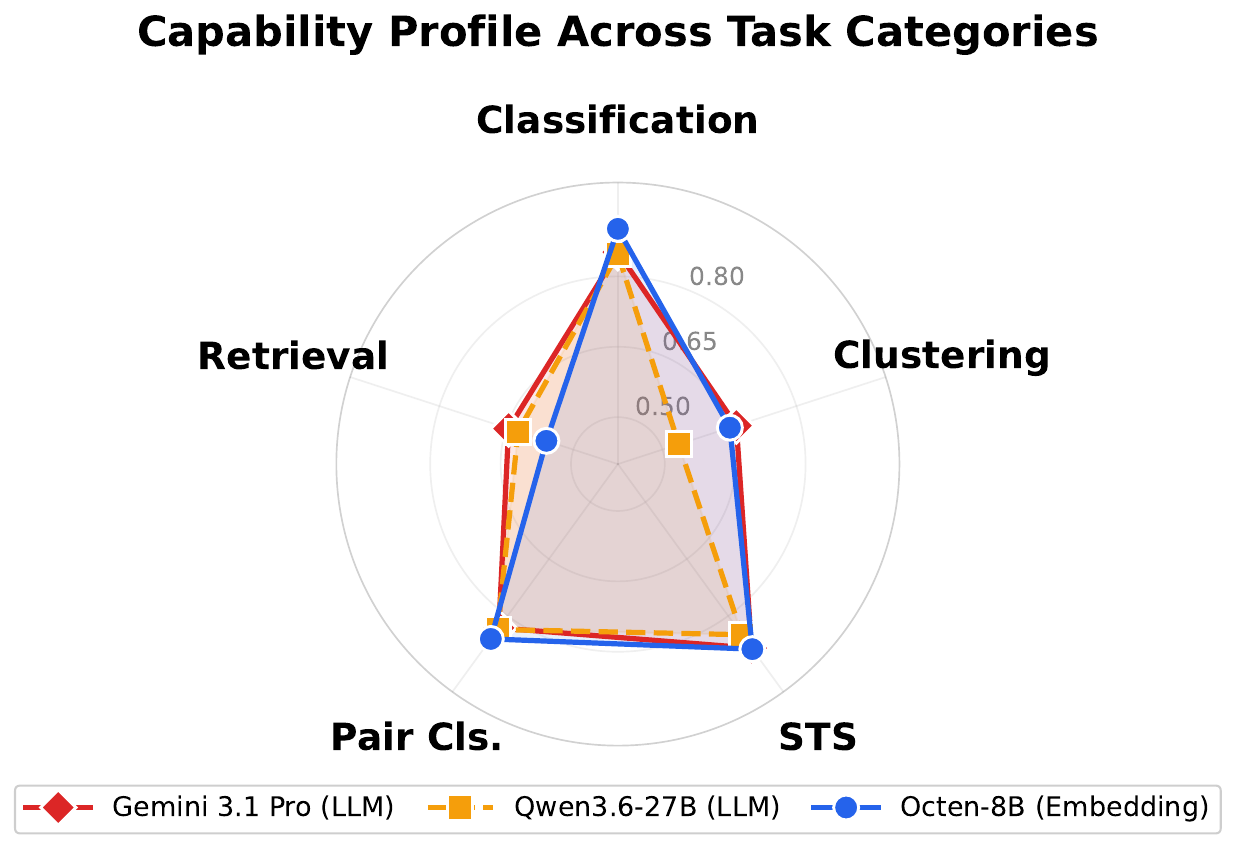}
\caption{\textbf{Capability profiles across five task categories.}
Gemini~3.1~Pro (red), Qwen3.6-27B (amber), and Octen-8B (blue).
The LLMs lead on retrieval, while Octen-8B leads on classification and remains competitive across categories.}
\label{fig:radar}
\end{figure}

\begin{figure}[htbp]
\centering
\includegraphics[width=\linewidth]{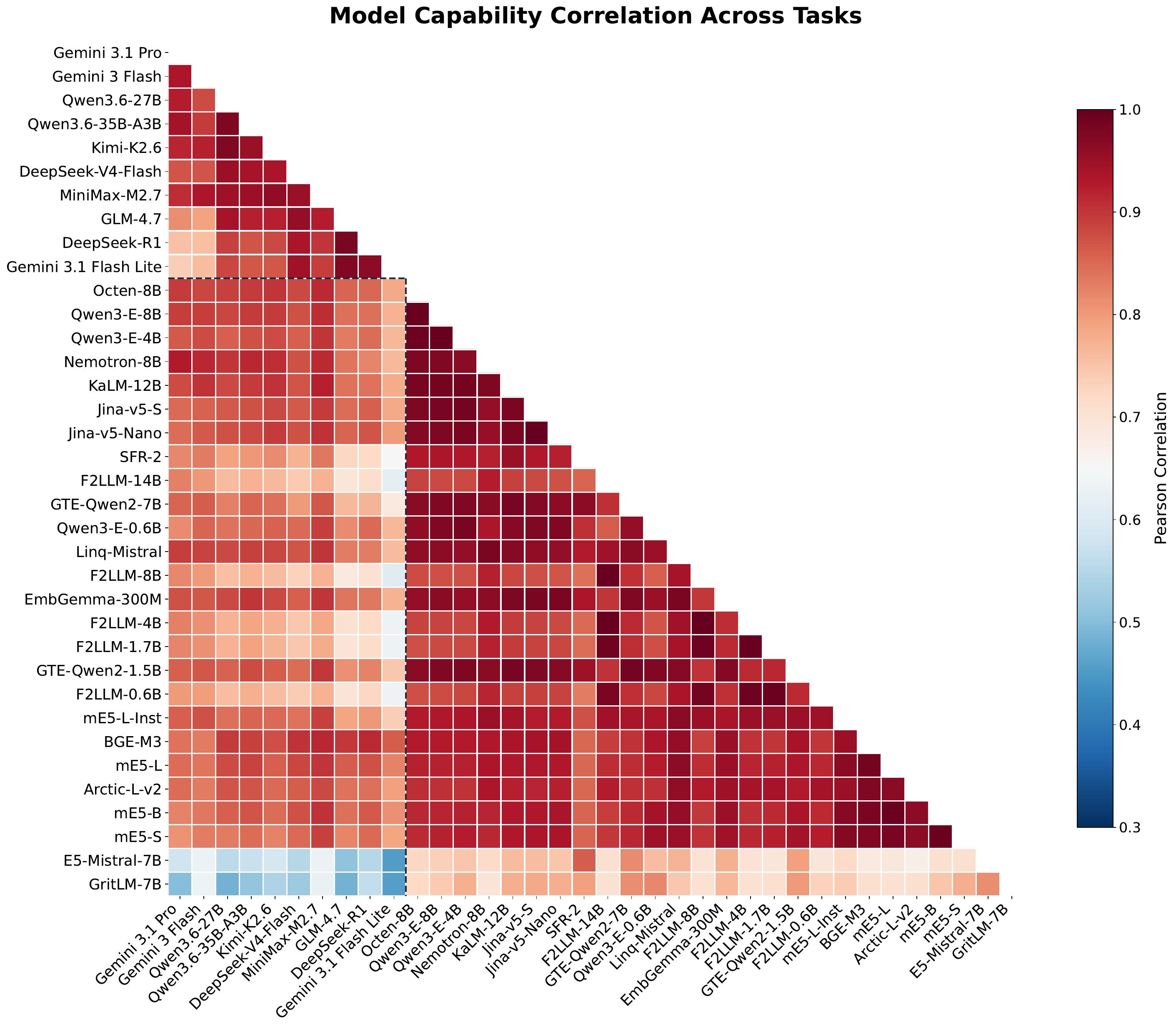}
\caption{\textbf{Cross-model task-behaviour correlation.}
Pearson correlations based on per-task \mteblm{} scores.
LLMs and embedding models form visually distinct clusters: LLMs are strongly intercorrelated while embedding models cluster by architecture family.
Lower inter-paradigm correlations are consistent with their different category profiles.}
\label{fig:heatmap}
\end{figure}

\FloatBarrier

\subsection{Per-Category Model Rankings}
\label{app:rankings_figs}

\begin{figure*}[htbp]
\centering
\includegraphics[width=\textwidth]{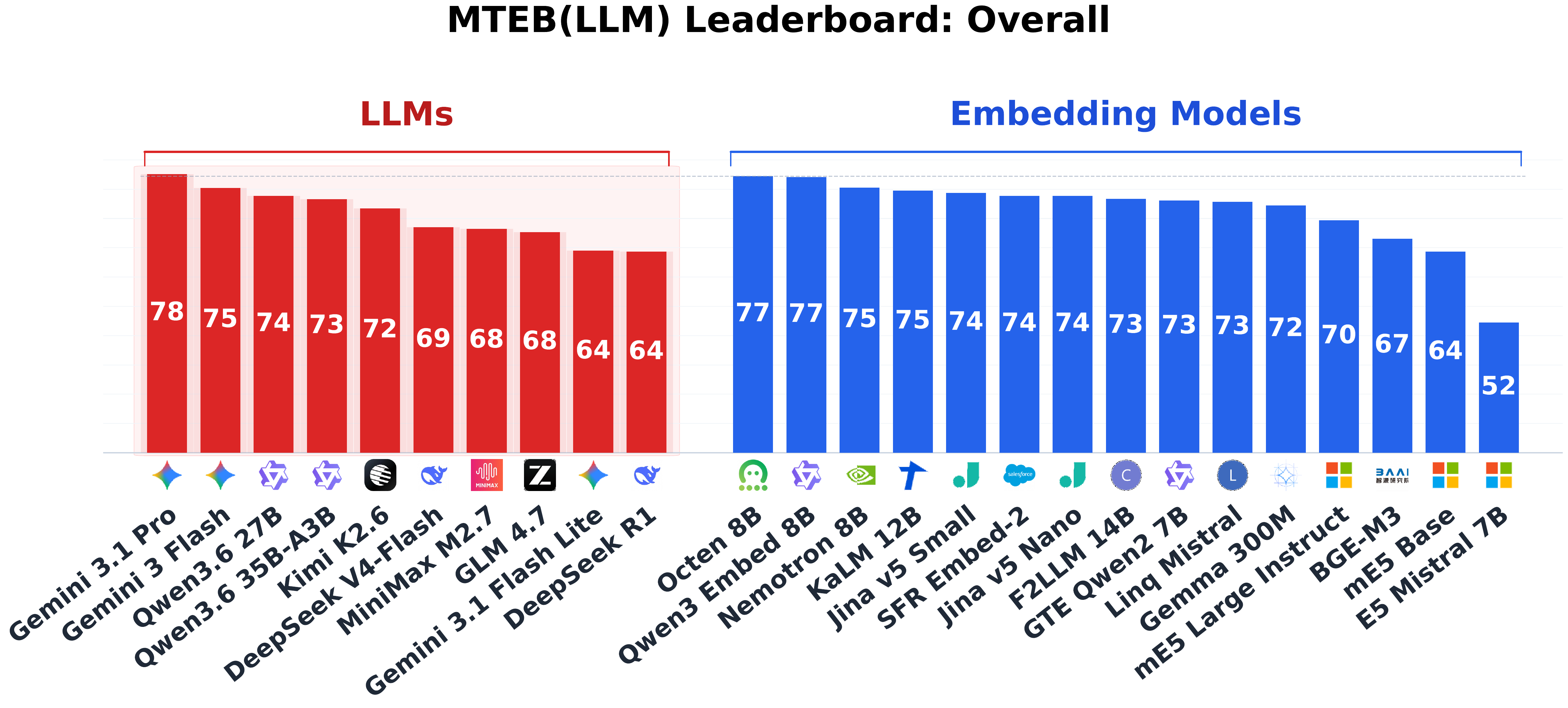}
\caption{\textbf{Overall rankings.} Pro (77.6) narrowly leads Octen-8B (77.2); the difference lies within statistical noise ($p$\,=\,0.85). Flash-Lite (64.5) ranks near the bottom.}
\label{fig:ranking_overall}
\end{figure*}

\begin{figure*}[htbp]
\centering
\includegraphics[width=\textwidth]{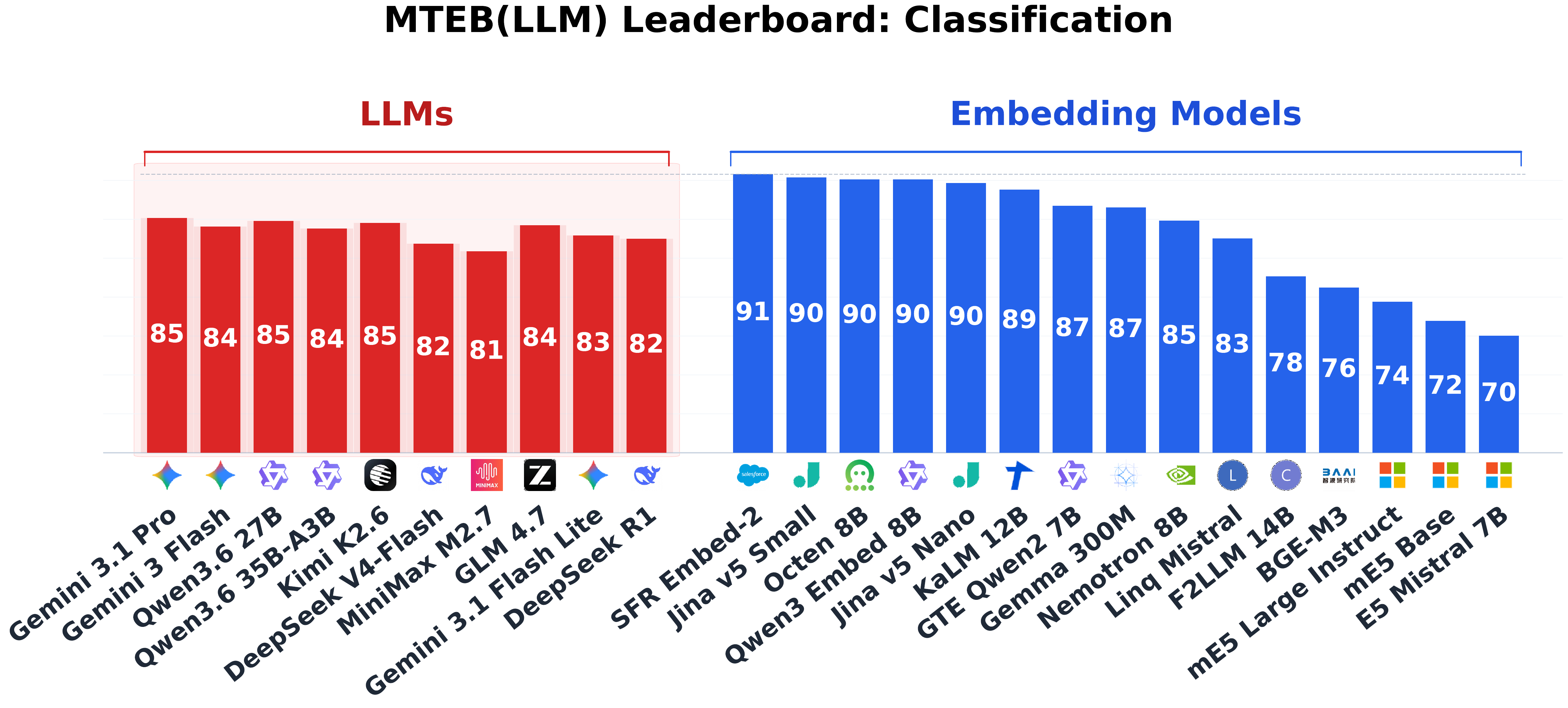}
\caption{\textbf{Classification rankings.} SFR-2 ranks first (90.8); Pro scores 85.2 and ranks below ten embedding models.}
\label{fig:ranking_clf}
\end{figure*}

\begin{figure*}[htbp]
\centering
\includegraphics[width=\textwidth]{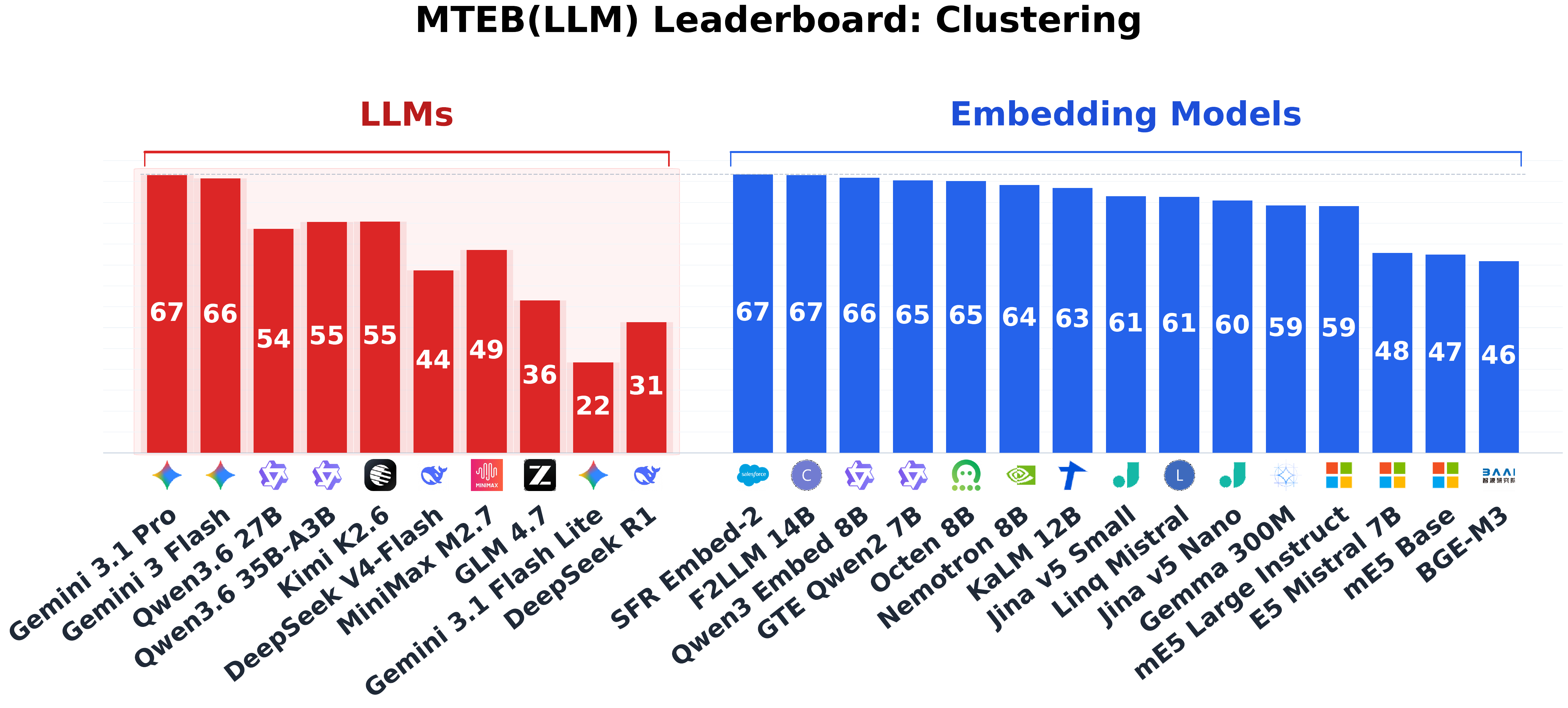}
\caption{\textbf{Clustering rankings.} Statistical tie; SFR-2 (66.7) edges Pro (66.6). Flash-Lite scores 21.7.}
\label{fig:ranking_clu}
\end{figure*}

\begin{figure*}[htbp]
\centering
\includegraphics[width=\textwidth]{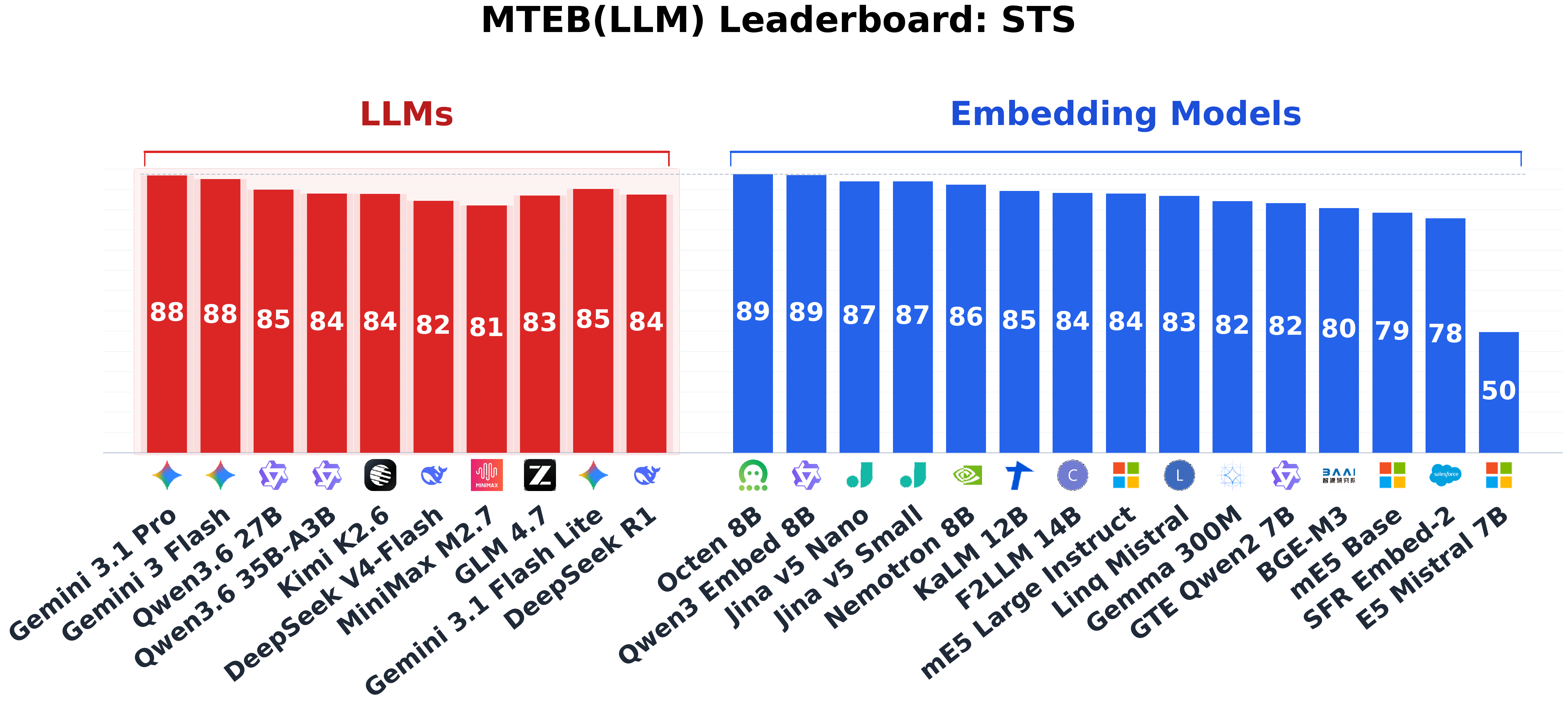}
\caption{\textbf{STS rankings.} Qwen3-E-4B (88.8) and Pro (88.5) are essentially tied; no significant paradigm difference ($p$\,=\,0.74).}
\label{fig:ranking_sts}
\end{figure*}

\begin{figure*}[htbp]
\centering
\includegraphics[width=\textwidth]{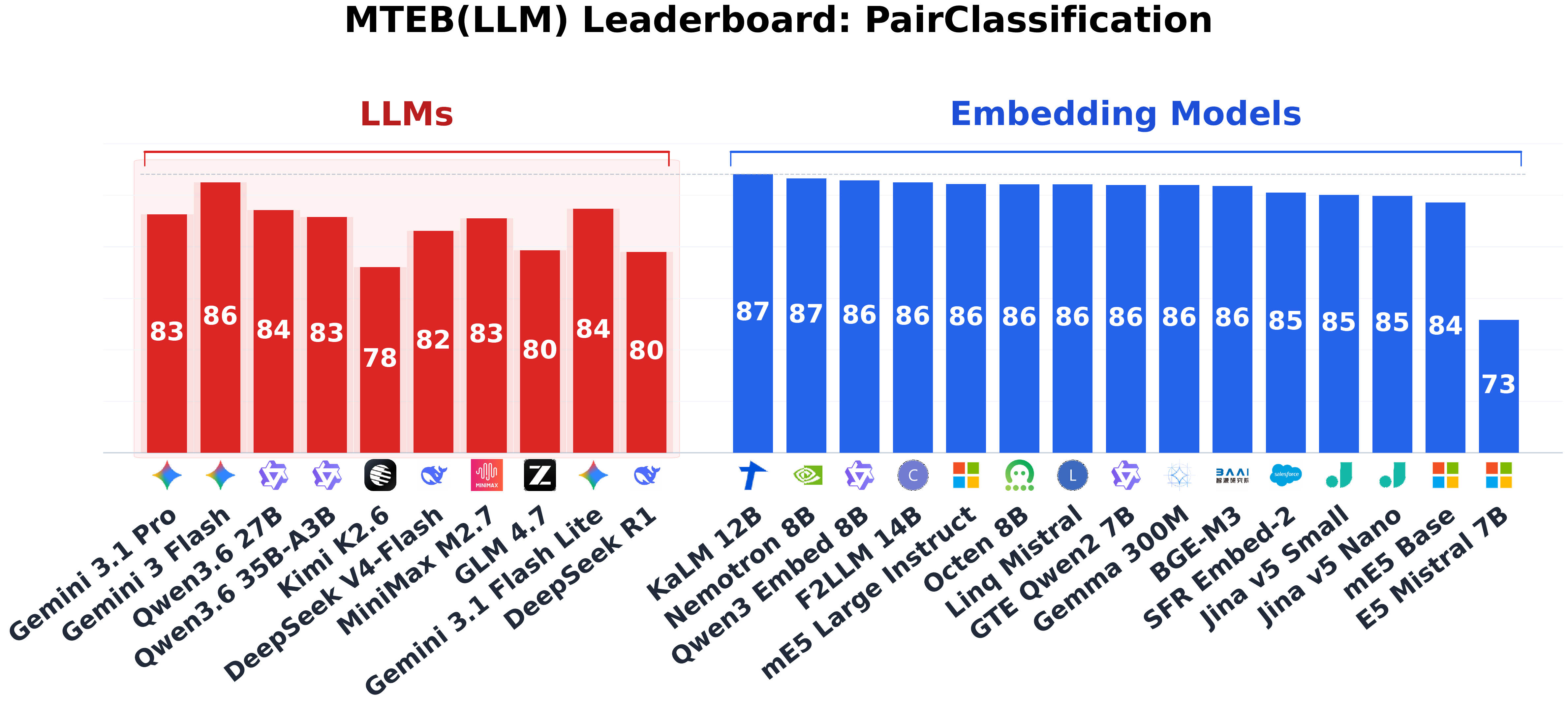}
\caption{\textbf{Pair Classification rankings.} KaLM-12B (87.1) outperforms Pro (83.2).}
\label{fig:ranking_pair}
\end{figure*}

\begin{figure*}[htbp]
\centering
\includegraphics[width=\textwidth]{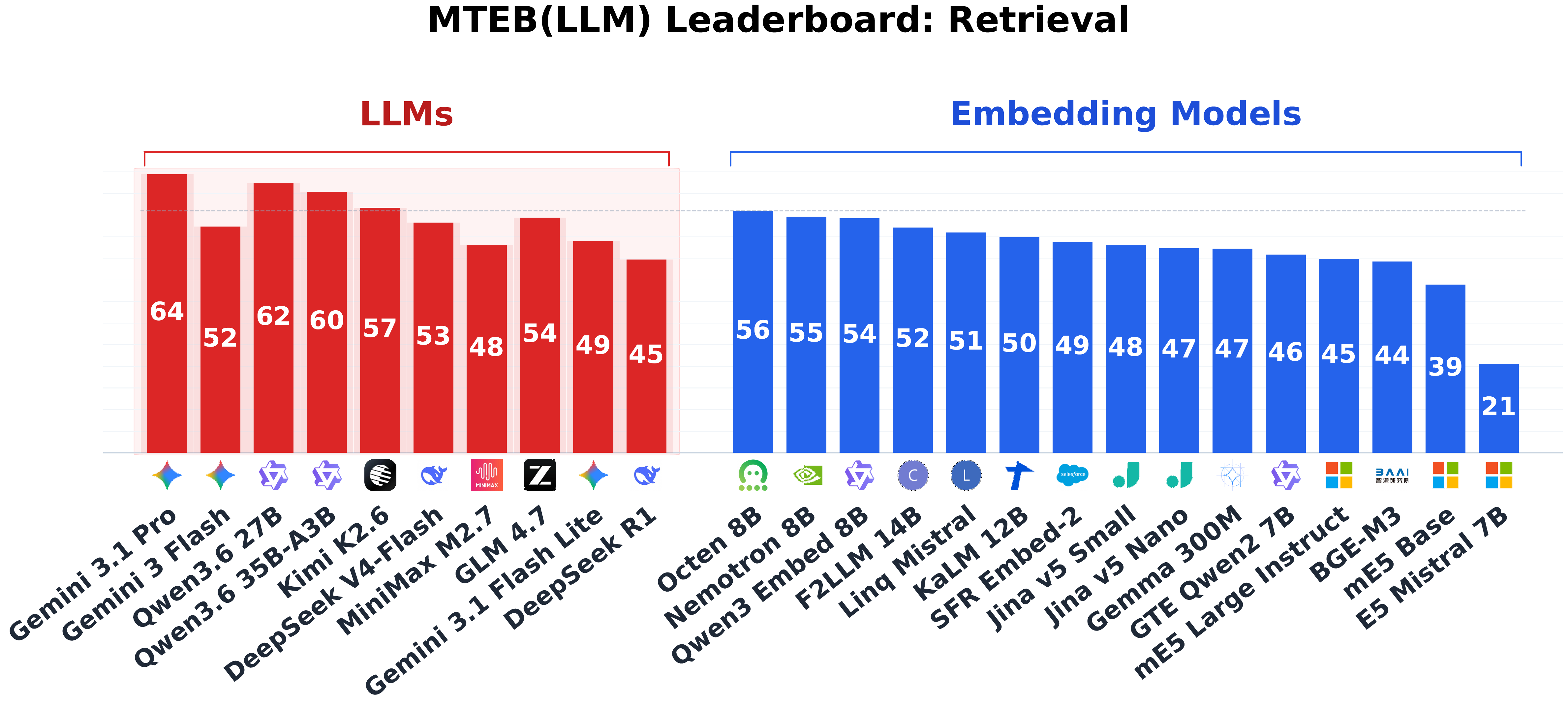}
\caption{\textbf{Retrieval rankings.} Pro leads the best embedding (64.5 vs.\ 56.0) and ranks first on five of six tasks.}
\label{fig:ranking_ret}
\end{figure*}

Figures~\ref{fig:ranking_overall}--\ref{fig:ranking_ret} rank all 36 evaluated models overall and within each of the five MTEB task categories.
They show category-dependent performance: LLMs lead on retrieval, embedding models lead on classification, and the two paradigms are statistically indistinguishable on clustering, STS, and pair classification.
Flash-Lite ranks near the bottom in every category, below the reasoning-capable Gemini models.

\FloatBarrier

\subsection{Full Retrieve-then-Rerank Matrix}
\label{app:rerank_matrix}

Table~\ref{tab:reranker_matrix} reports the complete matrix summarised by Figure~\ref{fig:rerank} (\S\ref{sec:rerank}): all four first-stage retrievers crossed with every cross-encoder and LLM listwise reranker, on BEIR and BRIGHT.

\begin{table*}[t]
\centering
\small
\setlength{\tabcolsep}{4pt}
\resizebox{\textwidth}{!}{%
\begin{tabular}{lcccccc}
\toprule
\textbf{First stage} & \textbf{Pure} & \textbf{BGE-Gemma} & \textbf{Qwen3-RR-4B} & \textbf{Qwen3-RR-8B} & \textbf{Qwen3.6-27B$^\dagger$} & \textbf{Qwen3.6-35B$^\dagger$} \\
\midrule
\multicolumn{7}{l}{\emph{BRIGHT (reasoning)}} \\
BM25 & 10.2 & 18.2 & 22.0 & 21.9 & \textbf{24.2} & 23.5 \\
BGE-large & 15.9 & 18.9 & 22.7 & 21.5 & \textbf{27.6} & 26.1 \\
GTE-MC-v1 & 15.9 & 19.2 & 23.9 & 23.0 & \textbf{28.8} & 27.9 \\
Qwen3-E-8B & 22.3 & 20.6 & 26.4 & 24.6 & \textbf{35.1} & 33.6 \\
\midrule
\multicolumn{7}{l}{\emph{BEIR (semantic)}} \\
BM25 & 39.8 & 50.5 & \textbf{53.2} & 52.5 & 52.1 & 51.1 \\
BGE-large & 52.7 & 55.3 & \textbf{58.6} & 57.6 & 57.6 & 56.2 \\
GTE-MC-v1 & 53.7 & 55.0 & \textbf{58.5} & 58.5 & 57.5 & 56.9 \\
Qwen3-E-8B & \textbf{63.1} & 56.8 & 60.3 & 59.5 & 58.9 & 58.9 \\
\bottomrule
\end{tabular}
}
\caption{\textbf{Retrieve-then-rerank matrix.} Average nDCG@10 for first-stage retrievers
crossed with cross-encoder and LLM listwise ($\dagger$) rerankers over 7 BRIGHT and 5 BEIR tasks. Bold = best per row.
An LLM reranker improves Qwen3-E-8B on BRIGHT (22.3$\to$35.1); on BEIR, the embedding alone scores highest (63.1).}
\label{tab:reranker_matrix}
\end{table*}

\section{Qualitative Error Analysis}
\label{app:samples}

We present model predictions from the evaluation files to illustrate differences between LLM and embedding approaches.
The examples cover classification ambiguity, STS calibration, retrieval precision, and differences between Pro and Flash.

\subsection{Classification: Over-Interpretation vs.\ Anchored Reference}

Several classification errors involve ambiguous inputs for which the LLM infers an intent that differs from the dataset convention.
With kNN, an embedding classifier can anchor its prediction to similar labelled examples, which helps on fine-grained tasks such as Banking77 and MassiveIntent.

\paragraph{Banking77 (77 classes): ``locate my card.''}
\begin{quote}
\small
\textbf{Input:} ``How do I locate my card?''

\textbf{Pro reasoning:} ``The user is asking how to locate their card, indicating that they have misplaced it or it is missing. This aligns with the intent for a lost or stolen card.''

\textbf{Pro prediction:} \texttt{lost\_or\_stolen\_card} \hfill \textcolor{red}{Incorrect (gold: \texttt{card\_arrival})}

\textbf{Flash prediction:} \texttt{card\_arrival} \hfill \textcolor{green!60!black}{Correct}
\end{quote}

\noindent Flash predicts the gold label; Pro predicts \texttt{lost\_or\_stolen\_card}.
The LLM infers ``locate'' implies loss; Flash's lower-depth reasoning follows the surface-level reading.
In the training data, ``locate my card'' aligns with delivery tracking.
The kNN classifier matches to examples like ``Where is my card?'' and correctly predicts \texttt{card\_arrival}.

\paragraph{Banking77: ``know when my card will arrive.''}
\begin{quote}
\small
\textbf{Input:} ``Is there a way to know when my card will arrive?''

\textbf{Pro reasoning:} ``The user is asking for information regarding the timeframe or expected date for their card's delivery, which aligns with the \texttt{card\_delivery\_estimate} category.''

\textbf{Pro prediction:} \texttt{card\_delivery\_estimate} \hfill \textcolor{red}{Incorrect (gold: \texttt{card\_arrival})}

\textbf{Flash prediction:} \texttt{card\_delivery\_estimate} \hfill \textcolor{red}{Incorrect (gold: \texttt{card\_arrival})}
\end{quote}

\noindent Both models make the same error here, illustrating the ambiguity between ``arrival'' and ``delivery estimate'' without labelled context.
The kNN classifier uses the full 77-class reference set to resolve such near-neighbour cases.

\paragraph{MassiveIntent (60 classes): ``remind me about my schedule.''}
\begin{quote}
\small
\textbf{Input (en):} ``remind me about my schedule for the afternoon''

\textbf{Pro reasoning:} ``The user is asking to be reminded about their schedule for the afternoon, which involves checking or querying their calendar events.''

\textbf{Pro prediction:} \texttt{calendar\_query} \hfill \textcolor{red}{Incorrect (gold: \texttt{calendar\_set})}
\end{quote}

\noindent ``Remind me'' is an action verb (set a reminder) but the LLM interprets it as information-seeking (query).
This example illustrates the ambiguity between action and query intents.

\paragraph{MassiveScenario (German): ``wir brauchen rosa.''}
\begin{quote}
\small
\textbf{Input (de):} ``wir brauchen rosa'' (we need pink)

\textbf{Pro reasoning:} ``The utterance 'wir brauchen rosa' (we need pink) implies adding an item ('rosa' / pink) to a shopping list or note-taking application.''

\textbf{Pro prediction:} \texttt{lists} \hfill \textcolor{red}{Incorrect (gold: \texttt{iot})}
\end{quote}

\noindent The LLM correctly reasons about the surface form but miscategorises the scenario: \texttt{iot} refers to controlling physical smart-home devices, not arbitrary colour references.

\begin{quote}
\small
\textbf{Input (de):} ``wie ist die einstellung f{\"u}r die helligkeit an meinem bildschirm'' (what is the brightness setting on my screen)

\textbf{Pro reasoning:} ``The user is asking about the brightness setting on their screen, which involves controlling or checking the state of a device. This falls under the IoT category.''

\textbf{Pro prediction:} \texttt{iot} \hfill \textcolor{red}{Incorrect (gold: \texttt{general})}
\end{quote}

\noindent Both errors reflect the fine-grained scenario taxonomy: \texttt{iot} refers to controlling physical smart-home devices, not arbitrary brightness settings or colour references.
The LLM correctly identifies device-related surface features but miscategorises the scenario.

\paragraph{AmazonCounterfactual: hypothetical ``would be great.''}
\begin{quote}
\small
\textbf{Input:} ``Would be great to have on my tablet instead of carrying the book.''

\textbf{Pro reasoning:} ``The review expresses a hypothetical situation ('have on my tablet') that has not actually happened, indicated by the phrase 'Would be great'. This describes an alternative condition and its potential consequence.''

\textbf{Pro prediction:} \texttt{counterfactual} \hfill \textcolor{red}{Incorrect (gold: \texttt{not-counterfactual})}
\end{quote}

\noindent The phrase ``would be'' supports the LLM's hypothetical reading. The dataset convention labels forward-looking wishes as \texttt{not-counterfactual}.
This may reflect a mismatch between a plausible linguistic interpretation and the dataset's annotation convention.

\paragraph{When LLMs succeed on classification.}
\begin{quote}
\small
\textbf{Input:} ``I still have not received my new card, I ordered over a week ago.''

\textbf{Pro prediction:} \texttt{card\_arrival} \hfill \textcolor{green!60!black}{Correct}
\end{quote}

\noindent In this less ambiguous example, the LLM matches the gold annotation.

\subsection{STS: Calibration and Semantic Precision}

The examples below show two kinds of disagreement between LLM scores and STS annotations: sensitivity to fine distinctions and higher scores for inferred mechanistic equivalence.

\paragraph{STSBenchmark: agreement on a simple paraphrase.}
\begin{quote}
\small
\textbf{S1:} ``open again, open anew'' \quad \textbf{S2:} ``open again or anew.''

\textbf{Pro:} 5.0 \quad \textbf{Gold:} 5.0 \quad \textbf{$|\Delta|$:} 0.0
\end{quote}

\noindent The LLM score matches the annotation for this near-identical paraphrase.

\paragraph{STSBenchmark: conservative scoring on ideological distinctions.}
\begin{quote}
\small
\textbf{S1:} ``\ldots a shift from a muslim identity to an islamist identity.''

\textbf{S2:} ``\ldots to change a muslim identity for an islamic identity.''

\textbf{Pro:} 4.5 \quad \textbf{Gold:} 5.0 \quad \textbf{$|\Delta|$:} 0.5
\end{quote}

\noindent The LLM distinguishes ``Islamist'' (political ideology) from ``Islamic'' (religious adjective); the gold standard treats them as equivalent.
This linguistically defensible distinction reduces agreement with the benchmark annotation.

\paragraph{BIOSSES: conservative scoring on biomedical paraphrases.}
\begin{quote}
\small
\textbf{S1:} ``It has recently been shown that Craf is essential for Kras G12D-induced NSCLC.''

\textbf{S2:} ``It has recently become evident that Craf is essential for the onset of Kras-driven non-small cell lung cancer.''

\textbf{Pro reasoning:} ``Both sentences state that Craf is essential for the development of Kras-driven NSCLC. Sentence 1 specifies the particular mutation (Kras G12D), whereas Sentence 2 is slightly more general.''

\textbf{Pro:} 3.8 \quad \textbf{Gold:} 4.0 \quad \textbf{$|\Delta|$:} 0.2
\end{quote}

\noindent The LLM correctly notes the specificity difference (``Kras G12D'' vs.\ ``Kras-driven'') and downweights accordingly.
The gold annotation gives 4.0, treating the two as highly similar; the LLM's 3.8 is slightly more conservative.

\paragraph{BIOSSES: LLM over-scores on mechanistic equivalence.}
\begin{quote}
\small
\textbf{S1:} ``More recently, IDH mutations and resultant 2-hydroxyglutarate (2HG) production in leukemia cells were found to impair TET2 function.''

\textbf{S2:} ``It has also been recently reported that mutations of the isocitrate dehydrogenase genes IDH1 and IDH2 inhibit TET catalytic activity.''

\textbf{Pro reasoning:} ``Both sentences describe the exact same biochemical pathway: mutations in IDH genes lead to the production of 2-hydroxyglutarate (2-HG), which in turn inhibits TET2 enzymatic activity.''

\textbf{Pro:} 3.8 \quad \textbf{Gold:} 3.2 \quad \textbf{$|\Delta|$:} 0.6
\end{quote}

\noindent The LLM identifies a mechanistic relationship and scores the pair 3.8, compared with the gold score of 3.2.
Expert adjudication would be needed to determine whether this difference reflects model calibration or annotation criteria.

\subsection{Retrieval: Reasoning Depth vs.\ Geometric Precision}

The retrieval examples show how joint query--document processing can help on reading-comprehension tasks and how broader associations can hurt precise statute retrieval.

\paragraph{FQuAD: reading comprehension in context.}
On French reading-comprehension retrieval, the LLM reads each candidate passage in the prompt and selects the one that answers the query, reaching Recall@1\,=\,92.0 versus 72.0 for the best embedding.
The LLM processes the query jointly with the candidate passages, which may explain part of the observed advantage.

\paragraph{AILAStatutes: where reasoning hurts.}
\begin{quote}
\small
\textbf{Query:} Legal case about a bank employee terminated for misconduct.

\textbf{Pro:} Returns 14 documents spanning broad legal concepts (contract law, employment law, due process). Gold: only 4 specific statutes.
The LLM's associative reasoning retrieves plausible but non-relevant provisions.

\textbf{Octen-8B:} Returns a narrower, more precise set via cosine similarity with legal embeddings.
\end{quote}

\noindent In this example, Pro retrieves plausible but irrelevant provisions, while the embedding model produces a narrower ranking.

\paragraph{TwitterHjerne: corpus-in-context precision.}
\begin{quote}
\small
\textbf{Query (Danish):} ``S{\o}nnen vil gerne lave \#pebern{\o}dder. De par gange jeg har pr{\o}vet det, blev de kun OK. Er der nogen, der kan anbefale en opskrift? \#twitterhjerne''
(``My son would like to make peppernuts. The few times I've tried, they were only OK. Can anyone recommend a recipe? \#twitterhjerne'')

\textbf{Pro:} Returns documents [1, 2]: exact match. \textbf{Recall@1\,=\,1.0 on this query.}

\textbf{Best embedding:} Recall@1\,=\,31.5 over the full task.
\end{quote}

\noindent For this query with explicit topical keywords, corpus-in-context retrieval returns the correct tweet first.
The overall task-level advantage for Pro (33.4 vs.\ 31.5 for embeddings) is modest, reflecting that both paradigms handle keyword-centric retrieval similarly.

\subsection{Model Scale Effects: Flash vs.\ Pro on the Same Input}

\paragraph{Banking77: Flash succeeds where Pro over-interprets.}

As shown above, Flash correctly predicts \texttt{card\_arrival} for ``How do I locate my card?'' while Pro predicts \texttt{lost\_or\_stolen\_card}.
In this example, more reasoning does not improve the prediction.
Pro's extended chain-of-thought introduces an interpretive step (``locate implies misplacement'') that Flash's lighter-weight reasoning bypasses.
Aggregate scores provide context: Pro remains slightly stronger than Flash on classification overall.

\paragraph{Retrieval scale effects.}

On FQuAD reading-comprehension retrieval, all three Gemini models outscore the best embedding, with a clear quality ladder:
\begin{itemize}
    \item Pro (Recall@1\,=\,92.0): reads and matches passages precisely.
    \item Flash (Recall@1\,=\,88.0): captures most of the advantage.
    \item Flash-Lite (Recall@1\,=\,84.0): still well above embeddings, even without reasoning.
    \item Best embedding (Recall@1\,=\,72.0): limited to single-vector matching.
\end{itemize}

Flash-Lite already outperforms the best embedding on this task; Pro adds a modest gain at much higher cost. This result is consistent with the reduced-thinking ablation (\S\ref{sec:ablation}).

\paragraph{Summary.}
The examples illustrate four patterns:
\begin{enumerate}
\item \textbf{Ambiguous label boundaries}: LLM predictions can differ from dataset conventions; kNN anchors decisions to labelled examples.
\item \textbf{STS calibration}: model scores can differ from annotations when they attend to different semantic details.
\item \textbf{Retrieval breadth}: joint processing can help reading comprehension but can also retrieve broadly related material.
\item \textbf{Model scale}: the value of additional capacity or reasoning varies across examples and must be weighed against cost.
\end{enumerate}
Estimating the frequency of these patterns requires a systematic error analysis.

\section{Extended Related Work}
\label{app:extended_related}

\subsection{LLMs as Text Encoders and Judges}

The boundary between generative LLMs and discriminative embedding models has blurred considerably.
LLM-as-a-judge frameworks \citep{zheng2023judging} repurpose frozen LLMs for evaluation. \citet{sun2023textclassificationlargelanguage} find that zero-shot LLM classification underperforms fine-tuned models on standard benchmarks and propose chain-of-thought prompting to narrow the gap.
Fine-tuned encoder models outperform zero-shot frontier LLMs across the classification benchmarks tested by \citet{bucher2024finetuned}, with margins of 5 to over 75 F1 points depending on task granularity. This is consistent with our 5.6-point advantage for kNN-equipped embedding pipelines.
The LOFT benchmark \citep{lee2024longcontextlanguagemodelssubsume} evaluates long-context LLMs on retrieval by placing entire corpora in the prompt (corpus-in-context prompting); our retrieval evaluation adopts the same protocol.

On the representational side, several lines of work extract embeddings from LLMs.
SGPT \citep{muennighoff2022sgptgptsentenceembeddings} uses weighted mean pooling over GPT decoder states. \citet{jiang2023scalingsentenceembeddingslarge} use in-context learning to extract sentence embeddings from decoder-only LLMs, while \citet{behnamghader2024llm2veclargelanguagemodels} adapt decoder-only LLMs into sentence encoders. \citet{springer2025repetitionimproveslanguagemodel} show that repeating the input can improve these embeddings by enabling effective bidirectional attention.
GritLM \citep{muennighoff2025generativerepresentationalinstructiontuning} goes further, unifying generation and representation in a single model that achieved strong performance on both MTEB and generative benchmarks at the time of its release.

These methods combine elements of both paradigms. At inference time, they produce fixed-length vectors compared via cosine similarity, so our cost accounting treats them as embedding models.
Several of our 26 evaluated models (e.g., E5-Mistral, GritLM) are built on LLM backbones and incorporate LLM-derived representations.

HUME \citep{assadi2025humemeasuringhumanmodelperformance} also benchmarks LLMs as annotators on MTEB tasks and finds that they fall short of human judgment on reranking. This complements our finding that zero-shot LLM classifiers underperform embedding pipelines with kNN access to labelled references.
Our study adds a controlled comparison on identical tasks, data, and metrics, together with exact cost accounting.

\subsection{Frontier LLMs and Open-Weight Reasoning Models}

LLMs have advanced rapidly since GPT-4 \citep{openai2024gpt4technicalreport}.
OpenAI's o1 \citep{openai2024o1} demonstrated extended chain-of-thought reasoning at inference time using reinforcement learning.
At the time of our evaluation, closed-source models with controllable reasoning included Gemini~3.1~Pro, GPT-5 \citep{singh2026openaigpt5card}, and Claude~Opus~4.6 \citep{anthropic2026claudeopus46}. These models support extended thinking tokens billed separately from output; newer models have since appeared (e.g.\ GPT-5.4).
Open-weight alternatives include Llama~3 \citep{grattafiori2024llama3herdmodels}, Qwen3 \citep{yang2025qwen3technicalreport}, and DeepSeek-R1 \citep{deepseekai2025deepseekr1incentivizingreasoningcapability}. DeepSeek-R1 shows that reinforcement-learning-based training can elicit strong reasoning at lower inference cost.
Gemma~3 \citep{gemmateam2025gemma3technicalreport} is a compact multilingual model family.

We evaluate ten LLMs across six families (\S\ref{sec:setup}), including DeepSeek-R1 and open-weight Qwen3.6, GLM, Kimi, and MiniMax models alongside Gemini~3. The cost and task-level patterns recur across this set, although their magnitude varies by model.
GPT-5 \citep{singh2026openaigpt5card} and Claude~Opus~4.6 \citep{anthropic2026claudeopus46} remain to be evaluated. The same evaluation framework can accommodate these and future models.

\subsection{Cross-Encoders and Late-Interaction Retrieval}

Cross-encoder rerankers (models that jointly encode query and document to produce a relevance score) and late-interaction models such as ColBERT \citep{khattab2020colbert} occupy a middle ground between bi-encoder speed and LLM reasoning depth.
We evaluate this middle ground directly (\S\ref{sec:rerank}) by crossing four first-stage retrievers with cross-encoder and LLM listwise rerankers on BEIR and BRIGHT \citep{sun2023rankgpt,weller2025rank1testtimecomputereranking}. Reranking improves reasoning-heavy retrieval, while the strongest embedding alone remains best on semantic retrieval.

\citet{li2024ragvslcllms} compare RAG with long-context LLMs on QA tasks. Long-context LLMs achieve higher quality at greater token cost, while a Self-Route hybrid retains 93--100\% of their quality at 38--61\% of the token cost, depending on the model. This result supports the hybrid recommendation in \S\ref{sec:discussion}.

\section{Extended Limitations}
\label{app:extended_limitations}

We expand on the six limitations summarised in \S\ref{sec:limitations}.

\paragraph{(i) LLM coverage.}
The ten evaluated LLMs span six families, dense and mixture-of-experts architectures, and reasoning and instruct variants. They provide a snapshot of a fast-moving frontier.
At current public rates, the Pareto frontier includes Gemini~3.1~Pro and the leading embeddings; the lowest-cost LLM evaluated scores below the best embedding.
Closed APIs not yet covered (GPT-5 \citep{singh2026openaigpt5card} and Claude~Opus~4.6 \citep{anthropic2026claudeopus46}) may have different cost--capability profiles, so the numerical conclusions should be updated as new evaluations become available.
The released evaluation pipeline makes this straightforward to extend as new models are released.

\paragraph{(ii) Supervision asymmetry.}
Classification uses labelled-reference kNN for embeddings and zero-shot prompting for LLMs.
This setup reflects deployments with scarce labelled data and gives the two pipelines different supervision. Five examples preserve performance on the small-label tasks and reduce Banking77 performance (\S\ref{sec:fewshot}).

\paragraph{(iii) Small-corpus retrieval.}
The corpus-in-context protocol places entire corpora (82--415 documents) in the LLM prompt, a setting that grants the LLM three structural advantages: the corpus fits entirely in context, prompt caching amortises input cost across queries, and the model can reason globally over the full document set.
Each is a property of the protocol rather than a measured effect; we did not vary corpus size, so we do not quantify how retrieval quality would change at larger scale.
At BEIR or production scale, placing the full corpus in context is generally impractical; an indexed first stage is needed, so the small-corpus cost gap is a lower bound for this protocol.
We complement corpus-in-context retrieval with bi-encoder, cross-encoder, and LLM-reranker comparisons on BEIR and BRIGHT (\S\ref{sec:rerank}). Reranking benefits reasoning-heavy retrieval; the strongest embedding retains the lead on semantic retrieval \citep{sun2023rankgpt,weller2025rank1testtimecomputereranking}.
nDCG@$k$ results for the small-corpus setting are available in the released data files and preserve the paradigm-level ordering.

\paragraph{(iv) Reasoning-control granularity.}
Reasoning controls differ by provider: Gemini's \texttt{reasoning\_effort=low} provides a soft hint, while the open models support fully disabled reasoning.
We use both settings. Reduced reasoning preserves or improves retrieval for four of six models across five families (Figure~\ref{fig:cost_breakdown}b); token accounting varies by provider.

\paragraph{(v) Throughput comparison.}
We serve two open-weight LLMs (Qwen3.6-27B and Qwen3.6-35B-A3B) and the embedding models on the same H100, removing API rate limits and hardware choice as confounds.
The resulting 2.5--736$\times$ gap is consistent with the difference between autoregressive decoding and a single encoder pass.

\paragraph{(vi) Task weighting sensitivity.}
Our aggregate metric weights all \mteblm{} tasks equally regardless of dataset size, task difficulty, or practical importance.
Alternative weighting schemes (by dataset size, normalised z-scores, or task-family weighting) may shift aggregate conclusions.
Aggregate numbers should therefore be interpreted alongside the reported category- and task-level results.
Alternative aggregation methods such as the Borda count used by MMTEB \citep{enevoldsen2025mmtebmassivemultilingualtext} are applicable to \mteblm{} and may produce different paradigm orderings.

\section{Cost and Throughput Details}
\label{app:tokens}

This section reports token usage, embedding throughput, cost sensitivity, and the cost methodology underlying \S\ref{sec:cost_results}--\ref{sec:throughput_results}.

\subsection{Detailed LLM Token Usage}
\label{app:tokens_table}

Table~\ref{tab:llm_tokens} provides the complete token-level breakdown for each LLM across all \mteblm{} tasks.
For each model, we report input, standard-output, cached, reasoning, and total tokens, together with the cost decomposition. Gemini reports reasoning separately, whereas OpenRouter includes it in the completion count.
The reasoning-heavy LLMs generate 8--26M reasoning tokens (Qwen3.6-27B 26M, GLM-4.7 25M, Kimi-K2.6 22M, Gemini~3~Flash 14M, Pro 9M), compared with 1--3M standard-output tokens. Reasoning accounts for 28--81\% of total inference cost across these models: 81\% for Qwen3.6-27B and 28\% for DeepSeek-V4-Flash.
The non-reasoning Flash-Lite costs \$6.85 and ranks below most embeddings.

\begin{table}[ht]
\centering
\small
\setlength{\tabcolsep}{4pt}
\begin{tabular}{lrrrrrrrr}
\toprule
& \multicolumn{4}{c}{\textbf{Tokens (M)}} & \multicolumn{4}{c}{\textbf{Cost (USD)}} \\
\cmidrule(lr){2-5} \cmidrule(lr){6-9}
\textbf{Model} & In$_{nc}$ & Cached & Out & Think & In$_{nc}$ & Cached & Out+Th & \textbf{Total} \\
\midrule
Gemini 3.1 Pro & 12.2 & 27.7 & 1.8 & 8.6 & \$24.39 & \$5.54 & \$124.21 & \textbf{\$154.14} \\
Gemini 3 Flash & 12.4 & 27.5 & 1.8 & 14.3 & \$6.19 & \$1.38 & \$48.30 & \textbf{\$55.87} \\
Qwen3.6-27B & 39.5 & 0.0 & 2.7 & 26.1 & \$11.46 & \$0.00 & \$91.97 & \textbf{\$103.43} \\
Qwen3.6-35B-A3B & 39.5 & 0.0 & 2.5 & 26.0 & \$5.53 & \$0.00 & \$28.52 & \textbf{\$34.06} \\
Kimi-K2.6 & 40.1 & 0.0 & 2.6 & 21.7 & \$27.43 & \$0.00 & \$83.27 & \textbf{\$110.70} \\
DeepSeek-V4-Flash & 16.8 & 22.6 & 2.1 & 4.5 & \$1.65 & \$0.22 & \$1.29 & \textbf{\$3.16} \\
MiniMax-M2.7 & 37.9 & 0.0 & 1.6 & 10.0 & \$10.58 & \$0.00 & \$13.93 & \textbf{\$24.52} \\
GLM-4.7 & 38.8 & 0.0 & 2.4 & 24.9 & \$15.54 & \$0.00 & \$47.77 & \textbf{\$63.31} \\
Gemini 3.1 Flash Lite & 16.5 & 26.9 & 1.4 & 0.0 & \$4.13 & \$0.67 & \$2.05 & \textbf{\$6.85} \\
DeepSeek-R1 & 39.6 & 0.0 & 1.2 & 10.6 & \$27.73 & \$0.00 & \$29.65 & \textbf{\$57.38} \\
\bottomrule
\end{tabular}
\caption{\textbf{Detailed LLM token usage and cost} across all \mteblm{} tasks.
In$_{nc}$ = non-cached input (cached billed at 10\% of input rate); Think = reasoning tokens
(billed at output rate); Out+Th combines standard output and thinking cost.}
\label{tab:llm_tokens}
\end{table}

\subsection{Embedding Throughput and Cost}
\label{app:emb_throughput}

Table~\ref{tab:embedding_throughput} reports the throughput (tokens/second) and derived cost per million tokens for each of the 26 embedding models, benchmarked on a single NVIDIA H100 80GB HBM3 GPU at sequence length 512 with maximum batch size.
Costs range from \$0.00016/MTok (mE5-small, 118M parameters, 4.3M tok/s) to \$0.047/MTok (F2LLM-v2-14B, 14.7K tok/s).
Total benchmark cost spans 291$\times$ across the embedding models, compared with a 1{,}431$\times$ gap between the best embedding and Gemini~3.1~Pro.

Each model's total benchmark cost uses its own tokenizer's token count (4.4--5.5M tokens across vocabularies), avoiding tokenizer mismatch.

\begin{table}[ht]
\centering
\small
\begin{tabular}{lcrrr}
\toprule
\textbf{Model} & \textbf{Params} & \textbf{Tok/s} & \textbf{\$/MTok} & \textbf{Score} \\
\midrule
mE5-S & 118M & 4,314,796 & 0.0002 & 63.1 \\
mE5-B & 278M & 1,910,005 & 0.0004 & 64.4 \\
mE5-L-Inst & 560M & 640,842 & 0.0011 & 69.7 \\
BGE-M3 & 568M & 640,425 & 0.0011 & 66.5 \\
Arctic-L-v2 & 568M & 640,405 & 0.0011 & 65.3 \\
mE5-L & 560M & 640,126 & 0.0011 & 65.9 \\
EmbGemma-300M & 308M & 374,988 & 0.0018 & 72.2 \\
Jina-v5-Nano & 212M & 327,790 & 0.0021 & 73.9 \\
F2LLM-0.6B & 596M & 189,653 & 0.0036 & 69.9 \\
Qwen3-E-0.6B & 596M & 175,692 & 0.0039 & 72.5 \\
GTE-Qwen2-1.5B & 1.5B & 133,567 & 0.0052 & 70.8 \\
F2LLM-1.7B & 1.7B & 106,406 & 0.0065 & 71.7 \\
Jina-v5-S & 596M & 91,658 & 0.0075 & 74.4 \\
Qwen3-E-4B & 4.0B & 46,080 & 0.0150 & 75.9 \\
F2LLM-4B & 4.0B & 40,157 & 0.0172 & 71.9 \\
GTE-Qwen2-7B & 7.1B & 35,419 & 0.0195 & 73.1 \\
Octen-8B & 7.6B & 29,313 & 0.0236 & 77.2 \\
Qwen3-E-8B & 7.6B & 29,262 & 0.0236 & 77.0 \\
SFR-2 & 7.1B & 28,196 & 0.0245 & 73.9 \\
Nemotron-8B & 7.5B & 27,808 & 0.0249 & 75.3 \\
GritLM-7B & 7.2B & 27,730 & 0.0249 & 51.2 \\
Linq-Mistral & 7.1B & 27,677 & 0.0250 & 72.8 \\
E5-Mistral-7B & 7.1B & 27,522 & 0.0251 & 52.2 \\
F2LLM-8B & 7.6B & 25,957 & 0.0266 & 72.2 \\
KaLM-12B & 11.8B & 19,244 & 0.0359 & 74.8 \\
F2LLM-14B & 14.0B & 14,665 & 0.0472 & 73.3 \\
\bottomrule
\end{tabular}
\caption{\textbf{Embedding throughput} on a single NVIDIA H100 80GB (median tokens/s over the
benchmark), per-MTok cost at \$2.49/hr spot, and mean (macro) \mteblm{} score.}
\label{tab:embedding_throughput}
\end{table}

\subsection{Cost Sensitivity Analysis}
\label{app:cost_sensitivity}

Table~\ref{tab:cost_sensitivity} shows how the LLM-to-embedding cost ratio varies under alternative hardware and pricing assumptions.
We consider five scenarios for embedding inference:
\begin{enumerate}
    \item \textbf{H100 spot} (\$2.49/hr, Lambda Labs, March 2026): baseline, 1{,}431$\times$
    \item \textbf{H100 on-demand} (\$3.99/hr): higher price, 893$\times$
    \item \textbf{A100 spot} (\$1.49/hr, est.\ 1.5$\times$ slower): 1{,}594$\times$
    \item \textbf{L4 spot} (\$0.49/hr, est.\ 3$\times$ slower): much lower throughput, 2{,}424$\times$
    \item \textbf{Commercial API} (e.g.\ \$0.10/MTok): fixed per-token pricing, 338$\times$
\end{enumerate}
In all five scenarios, the cost ratio exceeds 300$\times$.

\begin{table}[ht]
\centering
\small
\begin{tabular}{lrrr}
\toprule
\textbf{Hardware / Pricing Scenario} & \textbf{Emb.\ Cost} & \textbf{LLM Cost} & \textbf{Ratio} \\
\midrule
H100 spot \$2.49/hr (our setup) & \$0.108 & \$154.14 & 1,431$\times$ \\
H100 on-demand \$3.99/hr & \$0.173 & \$154.14 & 893$\times$ \\
A100 spot \$1.49/hr (est.\ 1.5$\times$ slower) & \$0.097 & \$154.14 & 1,594$\times$ \\
L4 spot \$0.49/hr (est.\ 3$\times$ slower) & \$0.064 & \$154.14 & 2,424$\times$ \\
Commercial API \$0.10/MTok & \$0.457 & \$154.14 & 338$\times$ \\
\bottomrule
\end{tabular}
\caption{\textbf{Cost sensitivity analysis.} LLM-to-embedding cost ratio under alternative hardware
and pricing scenarios. Costs compare Octen-8B with Gemini 3.1 Pro at fixed API pricing; ratios range from 338--2,424$\times$.}
\label{tab:cost_sensitivity}
\end{table}

\paragraph{Per-query economics and training amortisation.}
The benchmark-level ratios translate directly to per-query terms: both paradigms answer the same queries, so dividing by the query count rescales both sides equally and leaves the 29--1{,}431$\times$ range unchanged.
Two effects widen the gap further under sustained deployment.
First, embedding training is a one-time cost amortised over the deployment lifetime (and, for public checkpoints, across all users), so its per-query contribution vanishes at scale, whereas LLM inference cost recurs in full on every query.
Second, document embeddings are computed once and reused across queries, while corpus-in-context reading repeats per query and is only partially offset by prompt caching.
The benchmark-pass comparison is therefore conservative: it charges embeddings their full encoding cost while granting the LLM cached-input rates.

\subsection{Detailed Cost Methodology}
\label{app:cost_detail}

\paragraph{Embedding cost pipeline.}
For each model, we tokenize all \mteblm{} tasks with its HuggingFace \texttt{AutoTokenizer}. We then benchmark maximum-throughput processing on an NVIDIA H100 80GB at sequence length 512, using the largest batch that fits in memory. Cost is $(\text{tokens} / 10^6) \times (r_{\text{GPU}} / T)$, where $r_{\text{GPU}} = \$2.49$/hr (Lambda Labs spot pricing, March 2026) and $T$ is sustained throughput in tokens/hour.

Token counts range from 4.4M (Gemma-based tokenizers) to 5.5M (Mistral-based), a 26\% range driven by vocabulary differences.
Using a single proxy tokenizer (e.g., GPT-2 at 6.75M tokens) would overestimate embedding costs by 22--54\%.

\paragraph{LLM cost pipeline.}
Token usage is extracted from each provider's \texttt{usage\_stats} in every task's JSON result file.
For Gemini, \texttt{input\_tokens}\,=\,\texttt{prompt\_token\_count} (including cached tokens), \texttt{output\_tokens}\,=\,\texttt{candidates\_token\_count} (excluding reasoning), and \texttt{cached\_tokens}\,=\,\texttt{cached\_content\_token\_count}. Total tokens equal input plus output plus reasoning.
For the open models served via OpenRouter, the completion count already includes reasoning and \texttt{total}\,=\,input\,+\,output, with reasoning reported separately.
We bill every generated token, $\text{total}-\text{input}$, at the output rate. This equals output plus reasoning for Gemini and the reasoning-inclusive completion count for open models. Each model's reasoning share comes from its usage statistics (Figure~\ref{fig:cost_breakdown}).
For multilingual tasks (STS17, STS22v2, RTE3), per-language entries report different token counts and are summed; for classification tasks with multiple entries per split, all entries share identical \texttt{usage\_stats} and are counted once.

Gemini API prices per MTok input/output are \$0.25/\$1.50 for Flash-Lite, \$0.50/\$3.00 for Flash, and \$2.00/\$12.00 for Pro; cached input is billed at 10\%. Open models use OpenRouter public rates (March/June 2026).

\paragraph{Throughput measurement.}
We serve both paradigms on the same NVIDIA H100 and measure sustained tokens/second.
Open-weight LLMs (Qwen3.6-27B, Qwen3.6-35B-A3B) run under vLLM (BF16, tensor-parallel\,=\,1, 256 concurrent requests, 200/100 input/output tokens); embedding models run batch inference on the same GPU.
Holding hardware fixed removes API rate limits and GPU choice as explanations for the measured gap.

\section{Reproducibility}
\label{app:reproducibility}

All code, datasets, result files, and analysis scripts are released at \url{https://github.com/embeddings-benchmark/embedders-dilemma} and implemented with the MTEB framework.

\paragraph{LLM prompt templates.}
Prompt design follows a consistent structure across all five task categories:
\begin{itemize}
    \item \textbf{System prompt}: one sentence describing the task type and output format.
    \item \textbf{User prompt}: task-specific instruction, followed by the input text(s).
    \item \textbf{Output schema}: a Pydantic model with fields \texttt{reasoning} (free text, optional) and \texttt{output} (constrained to valid labels or numeric ranges). JSON mode is enabled for all calls.
\end{itemize}
Example system prompts: \emph{Classification}: ``You are a text classifier. Return a JSON with your reasoning and the predicted label from the provided list.'' \emph{Retrieval}: ``You are a retrieval assistant. Given a query and a set of numbered documents, identify the document IDs most relevant to the query. Return a JSON list of IDs in ranked order.''

\paragraph{Schema validation and failure handling.}
All LLM outputs are validated against the Pydantic schema after generation.
If validation fails (e.g., invalid label or malformed JSON), the prompt is retried with an appended correction instruction, up to twice per sample.
The final validation failure rate is $<$0.3\%; failed samples receive a null prediction and are counted as incorrect.

\paragraph{Token usage tracking.}
For each API call we log: \texttt{prompt\_token\_count}, \texttt{candidates\_token\_count}, \texttt{cached\_content\_token\_count}, \texttt{total\_token\_count}, and \texttt{thoughts\_token\_count} from \texttt{GenerateContentResponse.usage\_metadata}.
Thinking tokens are verified as $\texttt{total} - \texttt{prompt} - \texttt{candidates}$ to cross-check against \texttt{thoughts\_token\_count}; agreement was exact in all cases.
Pricing sources and throughput configurations are documented in \S\ref{app:cost_detail}.

\end{document}